\documentclass[10pt,twocolumn,letterpaper]{article}

\usepackage{iccv}
\usepackage{times}
\usepackage{epsfig}
\usepackage{graphicx}
\usepackage{amsmath}
\usepackage{amssymb}
\usepackage{pifont}
\usepackage{multirow}
\usepackage{cite}
\usepackage{adjustbox}
\usepackage{subcaption}
\usepackage{caption}

\usepackage[pagebackref=true,breaklinks=true,letterpaper=true,colorlinks,bookmarks=false]{hyperref}

\iccvfinalcopy 

\def\iccvPaperID{9309} 

\ificcvfinal\fi

\begin{document}

\title{Test Time Adaptation for Blind Image Quality Assessment}

\author{Subhadeep Roy$^\ast$ \quad Shankhanil Mitra$^\ast$ \quad Soma Biswas \quad Rajiv Soundararajan\\
Indian Institute of Science, Bengaluru, India\\
{\tt\small subhadeeproy2000@gmail.com, \{shankhanilm, somabiswas, rajivs\}@iisc.ac.in}}


\maketitle
\ificcvfinal\thispagestyle{empty}\fi

\def\thefootnote{$\ast$}\footnotetext{Authors contributed equally to this work}
\def\thefootnote{$\S$}\footnotetext{\url{https://github.com/subhadeeproy2000/TTA-IQA}}

\begin{abstract}
    While the design of blind image quality assessment (IQA) algorithms has improved significantly, the distribution shift between the training and testing scenarios often leads to a poor performance of these methods at inference time. This motivates the study of test time adaptation (TTA) techniques to improve their performance at inference time. Existing auxiliary tasks and loss functions used for TTA may not be relevant for quality-aware adaptation of the pre-trained model. In this work, we introduce two novel quality-relevant auxiliary tasks at the batch and sample levels to enable TTA for blind IQA. In particular, we introduce a group contrastive loss at the batch level and a relative rank loss at the sample level to make the model quality aware and adapt to the target data. Our experiments reveal that even using a small batch of images from the test distribution helps achieve significant improvement in performance by updating the batch normalization statistics of the source model. 
    
\end{abstract}

\section{Introduction}
The problem of image quality assessment (IQA) is extremely important in diverse image capture, processing, and sharing applications. However, a reference image is often not available for quality assessment. No reference (NR) or blind IQA primarily deals with the question of predicting image quality without using a reference image. Such NR IQA algorithms are often designed using machine learning approaches. More recently, deep learning based approaches have been extremely successful in achieving impressive performance. However, IQA applications are quite diverse and deal with several different distortions and distributional shifts. IQA models often have poor generalization ability and find it difficult to perform well under such shifts.  

Test time adaptation (TTA) has emerged as an important approach to address distributional shifts at test time \cite{c81}. It has been shown that by modifying a few global parameters of the model using a suitable loss that does not require the ground truth, one can significantly improve the performance of the model on the test data. Further, source-free adaptation, where the source data on which the original model was trained is not available while updating the model, is a realistic setting. While such approaches have been studied extensively in image classification literature \cite{ttt-rot,c13}, there is hardly any literature on TTA for IQA. 

There are multiple challenges in designing TTA for IQA. Typical losses used for TTA, such as entropy minimization \cite{c13}, are not applicable for IQA. For example, IQA is often studied in the regression context. This makes it difficult to extend models based on class confidences \cite{tta-conf} or class prototypes \cite{tta-proto} for classification to IQA. Also, the relevance of other self-supervised tasks such as rotation prediction \cite{c77}, context prediction \cite{c45}, colorization \cite{c47}, noise prediction \cite{c48}, feature clustering \cite{c49} for adapting IQA models is not clear. While contrastive learning has also been employed for TTA \cite{c37}, such a framework is not explicitly based on contrasting image quality, and its relevance is also not clear. 

Our main contribution is in the design of auxiliary tasks to enable TTA for IQA. We start with a source model trained on a large IQA  dataset and fine-tune the model on individual batches of test samples. The first task we introduce for adaptation is based on contrasting groups of low and high quality images in a batch.  Thus, we exploit the initial knowledge of the source model and try to adapt it by enforcing quality relationships among the batch of samples. Such a group contrastive (GC) learning approach fits naturally to our setting to account for any errors on individual samples that the source model may be prone to. 

In contrast to the GC learning that depends on the batch, our second auxiliary task is an image specific task based on distorted augmentations of different types. Here, our goal is to enable the model to rank the image quality of further distorted versions of each test sample. We explore the role of different distortion types to leverage the maximum benefit of this task. While GC learning is more effective when samples in a batch are diverse in quality, the rank order based learning is more effective when the quality of the images is not extremely poor. Thus, a combination of the tasks helps overcome the shortcoming of both tasks and leads to an overall superior performance.

We study the TTA problem under different settings of source and target datasets for multiple state of the art IQA models. Our results show significant improvements of the source model and the importance of TTA for IQA. We summarize our main contributions as follows:

\begin{itemize}
    \item We propose source-free test time adaptation techniques in the context of blind image quality assessment to mitigate distribution shifts between train and test data.
    \item We formulate two novel quality-aware self-supervised auxiliary tasks to adapt to the test data distribution. While group contrastive learning helps capture quality discriminative information among several images within a batch, rank ordering helps maintain the quality order between two different distorted versions of the same image.
    \item We show that our TTA method can significantly improve the performance of four different quality-aware source models, each on four different test IQA databases.
\end{itemize}

\begin{figure*}
\begin{center}
\includegraphics[width=0.95\textwidth,keepaspectratio]{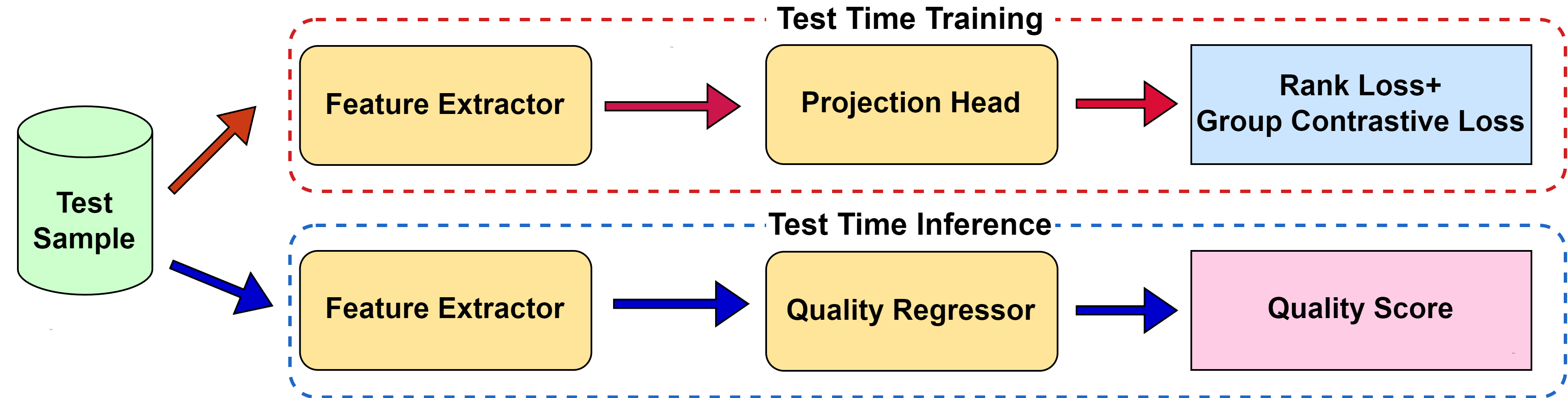}
\label{fig1}
\caption{\textbf{Block diagram of a general architecture for test time adaptation.} At test time training, the normalization layers of the feature extractor are adapted by optimizing the combination of rank and GC loss. At inference time, we predict the quality scores of test images using the updated feature extractor and pre-trained quality regressor. }
\end{center}
\end{figure*}

\section{Related Work}
\subsection{Test Time Adaptation}

One of the first pieces of work on TTA \cite{ttt-rot} introduces a joint training framework using a loss for the main task and a self-supervised auxiliary task loss. The choice of a relevant auxiliary task is a challenging part of the design. Sun \etal \cite{ttt-rot} use rotation prediction as a pre-text task for image classification applications. Researchers also explore simpler tasks that are highly correlated with the main task, such as feature alignment with the batch and a simple contrastive loss framework \cite{c37} for adaptation. 
However, these methods require the source data to train the source model all over again to enable TTA. 

TENT \cite{c13} studies source-free TTA, where entropy minimization-based adaptation even outperforms some  methods that use the source data. Moreover, only the batch normalization parameters of the model are adapted in TENT.  
There are several works on batch norm statistics adaptation \cite{c74,c39} to improve the robustness of the model at test time.
SHOT \cite{c40} presents a clustering-based pseudo-labeling
method to align features from the target domain to the source domain using an information
maximization loss. 
While a plethora of methods exists as above, the auxiliary tasks in these methods are not relevant for IQA. 

\subsection{No-Reference Image Quality Assessment}


Most classical NR IQA algorithms are mainly based on natural scene statistics (NSS) \cite{c16,c18,c19}. In \cite{c14}, the naturalness of the distorted image in the wavelet domain is modeled based on NSS. Saad \etal. \cite{saad} also design the NSS
model in the discrete cosine transform (DCT) domain. CORNIA \cite{c52}, and HOSA \cite{c53} are among the earliest codebook learning based methods to predict quality. With the emergence of deep learning and the availability of large subject-rated IQA databases, various general-purpose NR IQA methods have been designed based on convolutional neural network (CNN) architectures \cite{c55,c57,nrfr,c58}. Some of these methods require end-to-end training \cite{nrfr,c58} of deep neural networks (DNN), while others \cite{c61,c63,tres,c67} are based on updating pre-trained models with some modifications.  

 More recently, Zhang \etal \cite{c61} propose a method to train bilinear DNNs to simultaneously model both authentic and synthetic distortions. Su \etal \cite{c63} design a self-adaptive hyper network to provide weights for the quality prediction module parameters and handle various types of distortions and content in the images. Several methods \cite{tres,c67} explore transformer-based architectures along with a CNN to capture dependencies between local and global features. MetaIQA \cite{meta} proposes meta-learning on synthetic distortions by using a shared quality prior knowledge model to adapt to any kind of distortion.

All the existing methods assume that the train and test data come from the same distribution. If there is a distribution shift across different databases, we need adaptation of the pre-trained model to learn about the target distributional information.

\section{Methodology}
We propose a novel self-supervised \textbf{T}est \textbf{T}ime \textbf{A}daptation technique for \textbf{I}mage \textbf{Q}uality \textbf{A}ssessment (\textbf{TTA-IQA}) to adapt pre-trained quality models and mitigate distribution shifts between the source and target data. We consider source-free TTA, where we only have access to the pre-trained quality-aware models and no access to the source training data. 
When a batch of test data $D=\left\{x_j\right\}_{j=1}^{n}$ arrives, we adapt the model using the batch without knowledge of corresponding ground truth $\left\{y_j\right\}_{j=1}^{n}$.

\subsection{Approach}
Let the model trained on the source data be $f_\theta$ where $\theta=(\theta_e,\theta_c)$ corresponds to the parameters of the network, and $\theta_e$ and $\theta_c$ correspond to the parameters of the feature extractor and regression layers. Thus, for an input image $x$, we denote $f_\theta(x)=f_{\theta_c}(f_{\theta_e}(x))$. Since our goal is to learn the distribution shift between train and test data, we only update the parameters of the feature extractor, $\theta_e$, to align the features between the train and test distributions in a lower dimensional space.

The key challenge in TTA is the choice of a self-supervised auxiliary task that is highly correlated with the main task of IQA. However, relying too much on the auxiliary task can affect the performance of the main task. To prevent loss of learnt information induced by the auxiliary task \cite{simclr}, we first project the features to a lower dimensional space using a non-linear projection head $f_{\theta_s}$, parameterized by $\theta_s$. The auxiliary task drives the adaptation of the feature extractor through the projection head. Thus our model now has three parts parametrized by $(\theta_e,\theta_s,\theta_c)$ to resemble a Y-shape as shown in  Figure \ref{fig1}.

When a batch of test instances $D$ arrives,  we extract features from the feature extractor, followed by the projection head, and update the set of parameters ($\theta_e,\theta_s)$ by optimizing a self-supervised objective function $\mathcal{L}_s(D)$. Updating all the model parameters of the feature extractor $\theta_e$ can cause the model to diverge too much from training, and the performance can drop drastically. Inspired by prior work on test time entropy minimization \cite{c13} and improving robustness for test data \cite{c39}, we only update the linear and lower-dimensional feature modulation parameters. In a neural network, normalization layers satisfy these properties. So we only adapt the batch normalization layers by updating the affine parameters to mitigate distributional shifts. 
Thus we update these parameters by optimizing the auxiliary task loss given by 
\begin{equation}
\theta_e^*,\theta_s^* = \underset{{({\theta}_e,\theta_s)}}{\operatorname{argmin}} 
 \mathcal{L}_s(D)
\end{equation}
After optimizing the above loss function, 
we use the updated feature extractor parameters $\theta_e^*$ and pre-trained quality regressor parameters $\theta_c$ to predict the quality for a batch of test data. 
Since the distribution across batches can vary significantly, we ignore the earlier updated model and start the TTA based on source model weights for new test instants. Thus the updated target model always only depends on the incoming test data.
 
\begin{figure*}[h]
\begin{center}
\includegraphics[width=\textwidth]{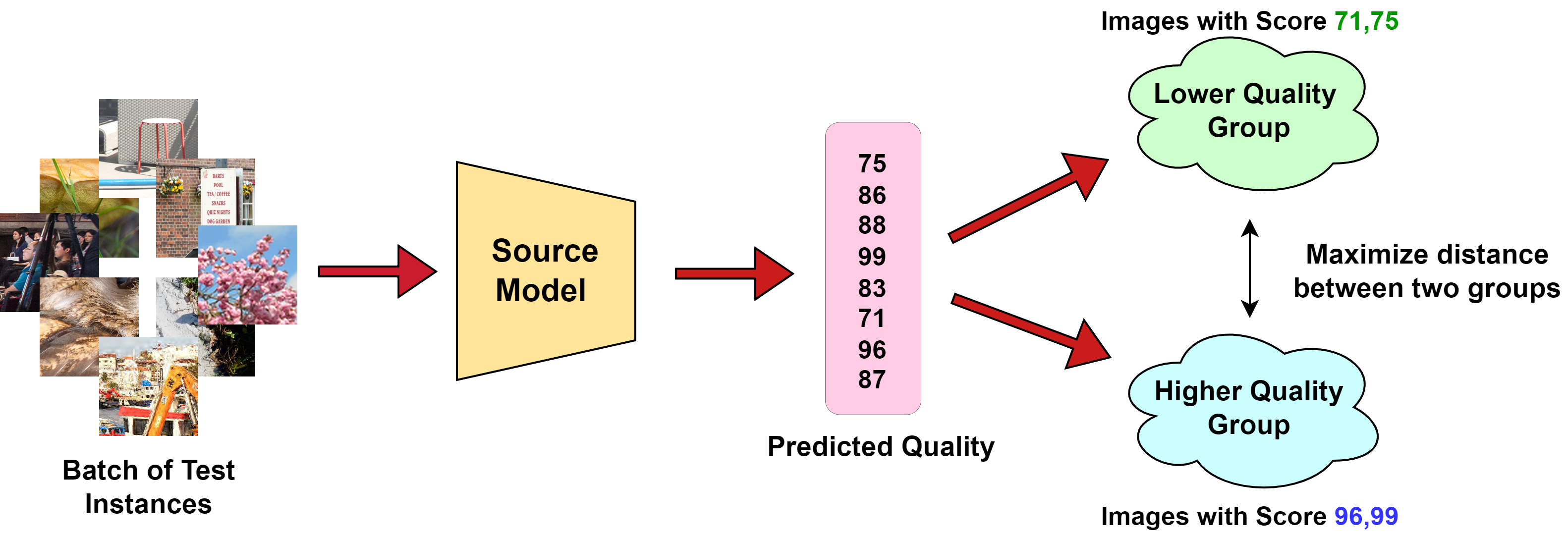}
\caption{\textbf{The overview framework of group contrastive loss for test time adaptation}. For a batch of images, two different groups are formed based on pseudo-labels given by the source model. The group contrastive loss tries to minimize the distance between features extracted from images belonging to the same group while maximizing the distance between features from different groups.}
\label{fig2}
\end{center}
\end{figure*}

\subsection{Self-Supervised Auxiliary Tasks}
Our goal is to carefully choose self-supervised auxiliary tasks that capture quality-aware distributional information to adapt the feature encoder. We formulate two novel and complementary self-supervised learning techniques which help to learn the distribution shift between train and test data. These self-supervised objective functions  are - 1) Group Contrastive Loss and 2) Rank Loss. While the GC loss works well when there is a reasonable separation of quality among the samples in a batch, the rank loss works better even when the quality of the batch samples is similar. On the other hand, the rank loss is meaningful only when the quality of the input image is not extremely low, while the GC loss is independent of the quality of a given image. Thus, a combination of the two losses renders our TTA extremely effective across various scenarios. 

\subsubsection{Group Contrastive Loss}

While contrastive learning has been used for TTA of deep image classification, its direct application does not appear relevant to the task of IQA. Thus, we introduce 
group contrastive (GC) learning as an auxiliary task for TTA of IQA models. 
In particular, we make two groups of images from a single batch of $N$ images based on the pseudo-labels given by the pre-trained source model. We sort the images in ascending order as $x_{(1)}$,$x_{(2)}$,\ldots,$x_{(N)}$ based on the pseudo-labels, where $x_{(i)}$, $i=1,2,\ldots,N$, corresponds to the $i^{th}$ lowest quality image in the batch.  We then segregate images with high quality scores (say, top $p$ fraction of the data in a batch) and include them in a group of higher-quality images. Similarly, we separate out lower quality images (say, the lowest $p$ fraction of the images in a batch) together to form another group. Here we assume that $pN$ and $(1-p)N$ are integers for simplicity; else, they can be rounded off to the nearest integers.

The premise behind our loss for GC learning is that images from the same quality group should give similar feature representations in a lower dimensional space while features of images from different groups are separated out. Thus image pairs from the same group act as positive pairs, and image pairs from different groups act as negative pairs. By separating out these two groups, the model adapts itself by better separating the intermediate quality samples.

Let a positive pair $x_{(i)}$ and $x_{(j)}$ come from the same group, i.e., either both $i,j\leq pN$ or $i,j > (1-p)N$. We use a modified NT-Xent contrastive loss \cite{simclr} as our objective function. For a  pair of images coming from the lower quality group where $i,j\leq pN$ and $i \neq j$, the GC loss is defined as 
\begin{equation}
    \mathcal{L}_{i, j}^{gc}=-\log \frac{\exp \left(\operatorname{sim}\left(\boldsymbol{z}_{(i)}, \boldsymbol{z}_{(j)}\right) / \tau\right)}{\sum_{k>(1-p)N}^N \exp \left(\operatorname{sim}\left(\boldsymbol{z}_{(i)}, \boldsymbol{z}_{(k)}\right) / \tau\right)},
\end{equation}
where $\mathbf{z}_{(i)} = f_{\theta_s}(f_{\theta_e}(x_{(i)}))$ represents the feature at the output of the projection head for sample $x_{(i)}$, $\operatorname{sim}$ refers to the cosine similarity between two features. 
and $\mathbf{z}_{(k)}$ is the feature extracted from an image from the higher quality group with $k > (1-p)N$. Also, $\tau$ represents temperature scaling parameter. While we define the loss above when $i,j\leq pN$, we can define a similar loss when $i,j>(1-p)N$. For all pairs of images within the same group, we obtain the GC loss and add them together to obtain the final loss, $\mathcal{L}^{gc}$ as
\begin{equation}
     \mathcal{L}^{gc}= \sum_{i=1}^{pN} \sum_{\substack{j=1\\ j \neq i}}^{pN} \mathcal{L}_{i, j}^{g c}+\sum_{i=(1-p) N+1}^{N} \sum_{\substack{j=(1-p)N+1\\j \neq i}}^{N} \mathcal{L}_{i, j}^{g c}.
\label{gc}
\end{equation}
A block diagram of the GC loss is shown in Figure \ref{fig2}. 

\begin{figure*}[h]
\begin{center}
\includegraphics[width=0.9\textwidth,keepaspectratio]{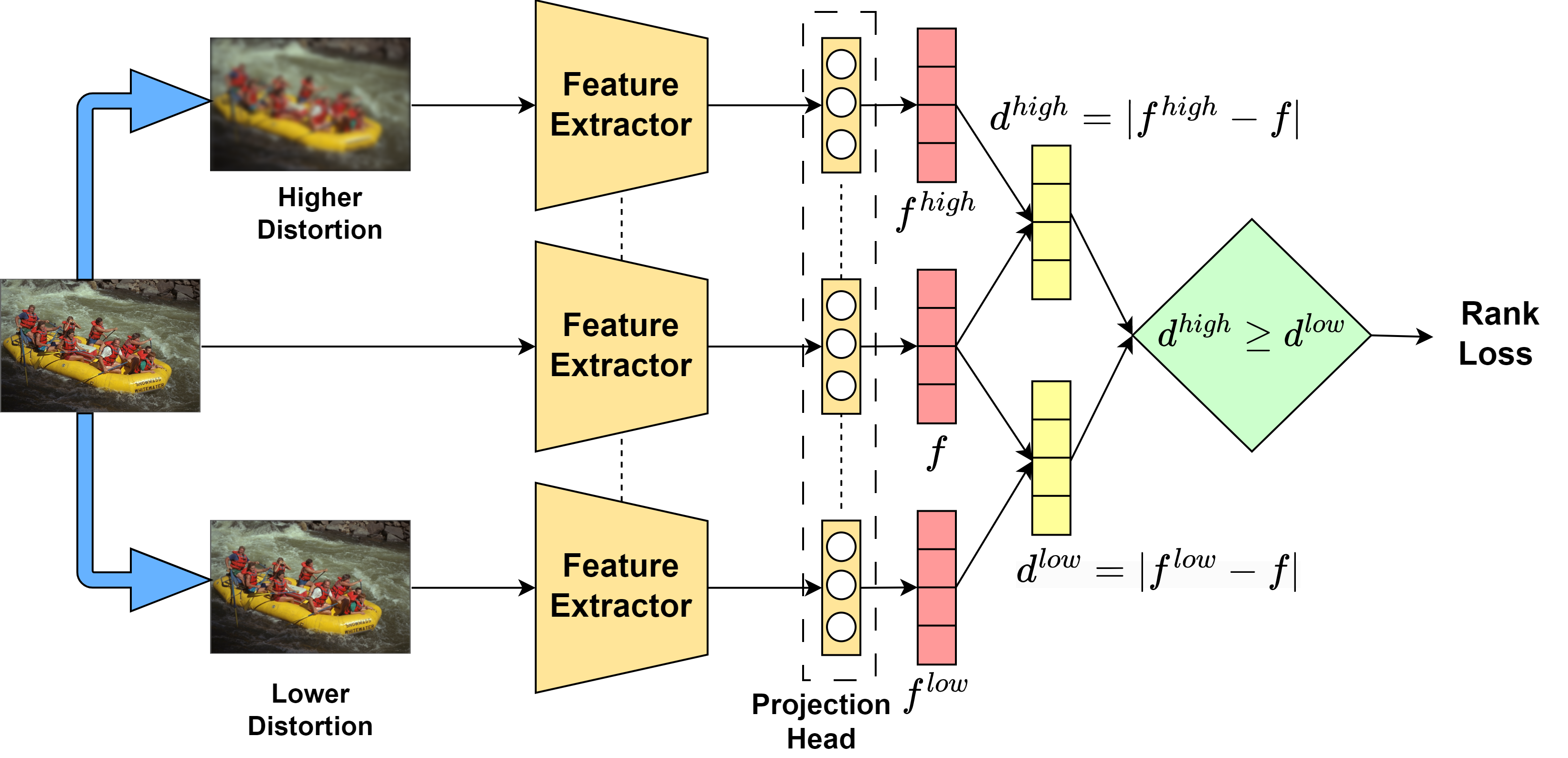}
\caption{\textbf{The rank loss framework for test time adaptation.} We distort the test image with two different degrees of degradation. We then project the triplet of images in the features space through the shared feature extractor and projection head. Distances calculated between the features extracted from each distorted image and the original test image are used to compare their rank order. }
\vspace{-8pt}
\label{fig3}
\end{center}
\end{figure*}

\subsubsection{Rank Loss}
\label{rank}
Given a sample of images from test data, the rank loss helps adapt the features by capturing quality-aware distributional information at the sample level. We introduce a ranking objective to learn the quality orders between two distorted versions of the test images that are quality discriminable. 
We distort each test image $x_i$, $i=1,2,\dots,N$, in the minibatch of size $N$ with two different degrees of degradation of a given distortion type. Let $x_i^{high}$ and $x_i^{low}$ denote the higher and lower distorted image, respectively. The degradation types include synthetic distortions such as blur, compression, and noise. We choose degradation levels randomly from two sets of parameters sufficiently farther apart.
Ideally, the distance between the features extracted from the test image and the highly distorted image should be greater than that between the features extracted from the test image and the lower distorted image. Our rank loss tries to capture this order and adapt the model to the test data.

We project the triplet of images $(x_i, x_i^{high}, x_i^{low})$ in the feature space through the feature extractor and the projection head. We measure the distance between the distortion-augmented image and the original test data using the Euclidean distance in the projected feature space.   
Mathematically, let $\mathbf{z}_i  = f_{\theta_s}(f_{\theta_e}(x_i))$ be the feature extracted for test sample $x_i$. Similarly we define $\mathbf{z}_i^{high}$ and $\mathbf{z}_i^{low}$.
Let $d_i^{high}$ denote the distance between $\mathbf{z}_i$ and $\mathbf{z}_i^{high}$. Similarly, $d_i^{low}$ denotes the distance between $\mathbf{z}_i$ and $\mathbf{z}_i^{low}$. The target model is now fine-tuned to obtain the correct ranking between the distances by achieving $d_i^{high} \geq d_i^{low}$ shown in Figure \ref{fig3}.

The probability of achieving this order is estimated by passing the difference in distances through a sigmoid function as
\begin{equation}
P_i = \operatorname{Pr}\left({d}_i^{high} \geq {d}_i^{low}\right) =\frac{\exp \left(d_i^{high}-d_i^{low}\right)}{1+\exp \left(d_i^{high}-d_i^{low}\right)}.
\label{eqn3}
\end{equation}
We use binary cross-entropy loss between true label $\Bar{P_i}=1$ and the predicted probability $P_i$ to obtain the rank loss as
\begin{equation}
\mathcal{L}_i^{r}\left(\bar{P_i},P_i\right)=-\bar{P_i} \log P_i-\left(1-\bar{P_i}\right) \log \left(1-P_i\right).
 \label{cross}
\end{equation}
For a batch of size $N$, the overall rank loss is given as
\begin{equation}
 \mathcal{L}^{r}= \sum_{i=1}^{N} \mathcal{L}_i^r.
\end{equation}

One of the challenges in the rank loss is selecting the distortion type for every image. The original test image may already be distorted, which poses difficulties in deciding the distortion type that can help TTA. 
For example, if the test image is extremely blurred and we blur it again with two different levels, both images may be visually indistinguishable, thereby limiting the extent to which adaptation can help. We exploit the source model's knowledge to overcome this limitation in choosing the distortion type. 
We predict the quality scores of the pair $(x_i^{high}, x_i^{low})$ for each distortion type using the source model. We hypothesize that the pair with the maximal difference in predicted quality is sufficiently different in visual quality. Thus choosing such a pair in Equation (\ref{cross}) can help adapt the model. 

Our overall self-supervised TTA objective function is a combination of both the rank and GC loss given as
\begin{equation}
\mathcal{L}_s=\mathcal{L}^{gc}+\lambda \mathcal{L}^r,
\label{eqn4}
\end{equation}
where $\lambda$ is a hyper-parameter used to combine the losses. 

\begin{table*}[h]
\centering
\begin{tabular}{|c|c|cc|cc|cc|cc|}
\hline
\multirow{2}{*}{Backbone} & Database & \multicolumn{2}{c|}{KonIQ-10k}          & \multicolumn{2}{c|}{PIPAL}          & \multicolumn{2}{c|}{CID2013}        & \multicolumn{2}{c|}{LIVE-IQA}       \\ \cline{2-10} 
                          & Method   & SROCC             & PLCC             & SROCC             & PLCC             & SROCC             & PLCC             & SROCC             & PLCC             \\ \hline
\multirow{3}{*}{TReS}     & Baseline & 0.6520          & 0.6955           & 0.3845          & 0.4078          & 0.5272          & 0.6463            & 0.5435          & 0.4450          \\
                          & Rotation & 0.6506          & 0.6805           & 0.4061          & 0.4114          & 0.5706          & 0.6651          & 0.5866          & 0.5311          \\
                          & TTA-IQA  & \textbf{0.6578} & \textbf{0.7074}     & \textbf{0.4278}  & \textbf{0.4204} & \textbf{0.6032} & \textbf{0.6710} & \textbf{0.6722} & \textbf{0.5963} \\ \hline
\multirow{3}{*}{MUSIQ}    & Baseline & 0.6304         & 0.6802           & 0.3190          & 0.3414           & 0.5173          & 0.6032          & 0.2596           & 0.3351          \\
                          & Rotation & 0.6577           & 0.7154          & 0.3665          & 0.3693          & 0.5487          & 0.6164          & 0.3512           & 0.3976           \\
                          & TTA-IQA  & \textbf{0.6693} & \textbf{0.7230} & \textbf{0.3743} & \textbf{0.3731} & \textbf{0.5499} & \textbf{0.6220}   & \textbf{0.3649} & \textbf{0.4031}  \\ \hline
\multirow{3}{*}{HyperIQA} & Baseline & 0.5861          & 0.6313          & 0.3037          & 0.3304          & 0.4895          & 0.6123          & 0.5143           & 0.4377          \\
                          & Rotation & \textbf{0.6033} & \textbf{0.6536} & 0.3268          & 0.3482          & 0.4902          & 0.6150          & \textbf{0.6268} & \textbf{0.5469} \\
                          & TTA-IQA  & 0.5960          & 0.6495          & \textbf{0.3653} & \textbf{0.3767}   & \textbf{0.5039}  & \textbf{0.5988} & 0.6218          & 0.5438          \\ \hline
\multirow{3}{*}{MetaIQA}  & Baseline & 0.5162          & 0.4460            & 0.3287          & 0.2955          & 0.7213          & 0.6817          & 0.7323          & 0.6732          \\
                          & Rotation & 0.5823          & 0.5311          & 0.3353          & 0.3042          & 0.7177          & 0.6745          & 0.7271          & 0.6868          \\
                          & TTA-IQA  & \textbf{0.5838} & \textbf{0.5428} & \textbf{0.4073} & \textbf{0.3510} & \textbf{0.7809}  & \textbf{0.7399} & \textbf{0.7999} & \textbf{0.7726} \\ \hline
\end{tabular}
\vspace{4pt}
\caption{Comparison of TTA-IQA with popular NR IQA methods and one popular auxiliary task - rotation prediction on authentically and synthetically distorted datasets. Bold entries imply the best performance for every individual quality-aware model on respective datasets. }
\label{t1}
\end{table*}

\section{Experiments}
\subsection{Quality Models, Datasets and Metrics}

We evaluate TTA on four popular IQA databases using four different quality-aware models. In particular, we consider state-of-the-art deep IQA models such as TReS \cite{tres}, MUSIQ \cite{c67} , HyperIQA \cite{c63}, and MetaIQA \cite{meta}. These models contain a ResNet backbone with batch normalization layers, which we model as the feature extractor $f_{\theta_e}$ and only update its batch normalization parameters. We model the rest of the network as the quality regressor part, $f_{\theta_c}$. TReS and MUSIQ use transformers as a part of their architecture, which we include as a part of the quality regressor in all our main experiments. We also explore the adaptation of transformers by including them as part of the feature extractor and updating their layer normalization statistics in the supplementary material.

We project the features extracted from the last layer of ResNet through a 256-dimensional fully connected (FC) layer corresponding to the self-supervised projection head $f_{\theta_s}$. These lower dimensional features are used to adapt the model at test time.
Three IQA models, TReS, MUSIQ, and HyperIQA are trained on the LIVEFB database \cite{c23} containing 39,810 images. MetaIQA is trained on  two synthetically distorted databases, TID2013 \cite{tid} and KADID-10k \cite{kadid}.

We choose challenging databases to evaluate the generalization capability of our TTA-IQA method. The test datasets are described as follows:

\textbf{KonIQ-10k} \cite{c24} is a popular in-the-wild authentically distorted database consisting of 10,073 quality-scored images.

\textbf{PIPAL} \cite{c25} is a large-scale IQA database for evaluating perceptual image restoration. It contains 29k distorted images, including 19
different GAN-based algorithms. 

\textbf{CID2013} \cite{c26} consists of 480 images in six image sets captured by 79 imaging devices.

\textbf{LIVE-IQA} \cite{LIVE} consists of 779 synthetically distorted images from 29 reference images.

We evaluate our results using Spearman’s rank-order correlation coefficient (SROCC) and Pearson’s linear correlation coefficient (PLCC).

\subsection{Implementation Details}
We implement our setup in PyTorch and conduct all the experiments with an 16 GB NVIDIA RTX A4000 GPU. During TTA, we randomly select a patch of size $224\times224$ from the input image and apply quality preserving augmentations such as horizontal flip and vertical flip before passing it through the network for TTA. We use the ADAM \cite{c72} optimizer with a learning rate of 0.001 and set the number of iteration as 3. All the experiments were run on five different seeds using a batch size of 8, and the final results were obtained by averaging.

With regard to the self-supervised auxiliary tasks, we use a combination of rank loss and GC loss as our objective function using $\lambda=1$ given in Equation (\ref{eqn4}). We choose $p=0.25$ to determine the groups. We calculate the GC loss using $\tau=1$. For the rank loss, the test image is synthetically blurred using a Gaussian blur filter of size $5\times5$ with two sets of standard deviations $\sigma$, for the Gaussian blur kernel. We keep $\sigma \in [40,80]$ for highly distorted images and $\sigma\in [1,20]$ for less blurred images. For compression distortions, we specify the  quality factor in $[80,95]$ for lower compression and $[30,60]$ for higher compression rates. Similarly, we add zero mean Gaussian white noise to the test image with variance in $[0.05,0.1]$ for higher distortion and in $[0.005,0.01]$ for lower distortion.

\begin{table*}[t]
\centering
\begin{tabular}{|c|cc|cc|cc|cc|cc|}
\hline
\multirow{2}{*}{Backbone} & \multicolumn{2}{c|}{Database}  & \multicolumn{2}{c|}{KonIQ-10k}          & \multicolumn{2}{c|}{PIPAL}          & \multicolumn{2}{c|}{CID2013}        & \multicolumn{2}{c|}{LIVE-IQA}       \\ \cline{2-11} 
                          & \multicolumn{1}{c|}{Rank} & GC & SRCC             & PLCC             & SRCC             & PLCC             & SRCC             & PLCC             & SRCC             & PLCC             \\ \hline
\multirow{3}{*}{TReS}    
                          & \multicolumn{1}{c|}{$\checkmark$}    & $\times$  & 0.6562          & 0.6989          & 0.4171          & 0.4204          & 0.6016          & \textbf{0.6736} & 0.6705          & 0.5908           \\
                          & \multicolumn{1}{c|}{$\times$}    & $\checkmark$  & 0.6516           & 0.6946          & \textbf{0.4666} & \textbf{0.5183} & 0.5366          & 0.6493           & \textbf{0.7160}   & \textbf{0.6290} \\
                          & \multicolumn{1}{c|}{$\checkmark$}    & $\checkmark$  & \textbf{0.6578} & \textbf{0.7074} & 0.4278           & 0.4204          & \textbf{0.6032} & 0.6710          & 0.6722          & 0.5963          \\ \hline
\multirow{3}{*}{MUSIQ}   
                          & \multicolumn{1}{c|}{$\checkmark$}    & $\times$  & 0.6549          & 0.7149          & 0.3768          & 0.3718          & 0.5216           & 0.6034          & 0.3634            & 0.4024          \\
                          & \multicolumn{1}{c|}{$\times$}    & $\checkmark$  & 0.6611          & 0.7176          & 0.3585          & 0.3642          & 0.5446           & 0.6159          & 0.3299          & 0.3914          \\
                          & \multicolumn{1}{c|}{$\checkmark$}    & $\checkmark$  & \textbf{0.6693} & \textbf{0.7230} & \textbf{0.3743} & \textbf{0.3731} & \textbf{0.5499} & \textbf{0.6220}   & \textbf{0.3649} & \textbf{0.4031}  \\ \hline
\multirow{3}{*}{HyperIQA}
                          & \multicolumn{1}{c|}{$\checkmark$}    & $\times$  & 0.5928          & 0.6455           & 0.3616          & 0.3732            & 0.5039          & 0.5991          & 0.5505          & 0.5050          \\
                          & \multicolumn{1}{c|}{$\times$}    & $\checkmark$  & \textbf{0.6094} & \textbf{0.6567} & 0.3333          & 0.3552          & \textbf{0.5120} & \textbf{0.6255} & \textbf{0.6331} & \textbf{0.5592}  \\
                          & \multicolumn{1}{c|}{$\checkmark$}    & $\checkmark$  & 0.5960          & 0.6495          & \textbf{0.3653} & \textbf{0.3767} & 0.5039           & 0.5988          & 0.6218          & 0.5438          \\ \hline
\multirow{3}{*}{MetaIQA} 
                          & \multicolumn{1}{c|}{$\checkmark$}    & $\times$  & 0.5580            & 0.5118          & 0.3992          & 0.3437          & 0.7579          & 0.7282          & 0.7894          & 0.7534          \\
                          & \multicolumn{1}{c|}{$\times$}    & $\checkmark$  & 0.5414          & 0.4636          & 0.3710          & 0.3287          & \textbf{0.7861} & \textbf{0.7067} & 0.7566         & 0.6927          \\
                          & \multicolumn{1}{c|}{$\checkmark$}    & $\checkmark$  & \textbf{0.5838} & \textbf{0.5428} & \textbf{0.4073} & \textbf{0.3510} & 0.7809           & 0.7399          & \textbf{0.7999} & \textbf{0.7726} \\ \hline
\end{tabular}
\vspace{4pt}
\caption{Ablation study results on authentically and synthetically distorted datasets using popular quality aware source models. Bold entries imply the best performance among all three settings.}
\label{t2}
\end{table*}

\subsection{Performance Evaluation}
Table \ref{t1} shows the performance of our method on all four test datasets using all four quality-aware models. In addition to the source model, we also compare with rotation prediction \cite{ttt-rot} as the auxiliary task during TTA. A comparison with such a task helps understand the role of quality-aware losses for the TTA of IQA models.

We observe that TTA-IQA using the combination of rank and GC loss outperforms the source models in all the cases.  
Note that the PIPAL dataset has a huge distribution shift from the authentically distorted LIVEFB dataset on which three of the models were trained. While most source models perform very poorly on PIPAL, we achieve 10\%-20\% improvement over the source models.
On the KonIQ-10k dataset, TTA-IQA gives around 1\%-10\%  improvement over the source model. As both KonIQ-10k and LIVEFB are authentically distorted datasets, the distribution shift is reasonably small, leading to smaller improvements using adaptation. 
CID2013 is also an authentically distorted dataset and gives similar improvements of around 2\%-13\%  over the source models.
Our experiments on the LIVE-IQA database provide a significantly greater improvement of 10\%-33\% owing to the shift from authentic to synthetic distortions for three models. 

\begin{figure*}[t]
\minipage{0.32\textwidth}
  \includegraphics[width=\linewidth]{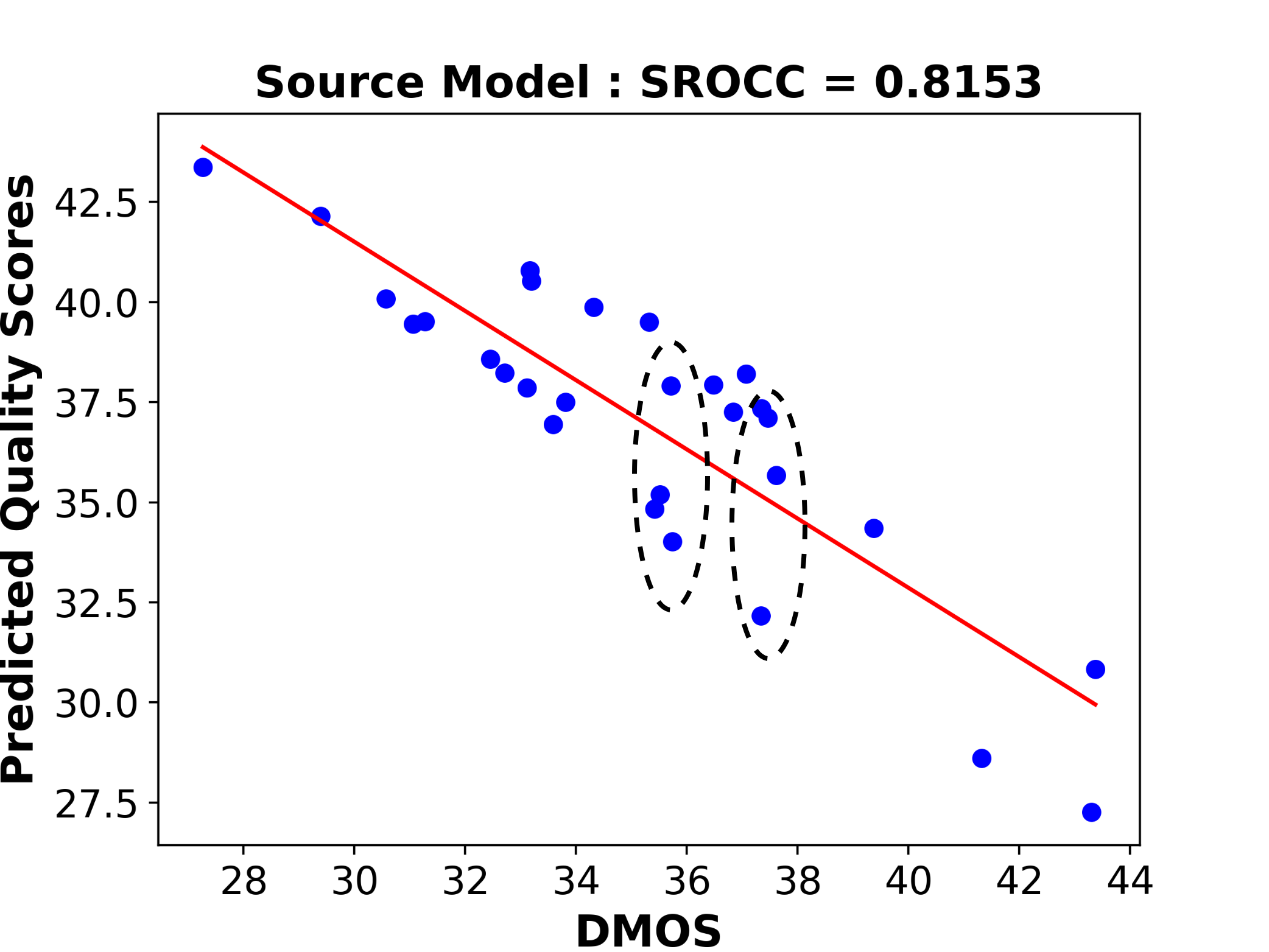}
\endminipage\hfill
\minipage{0.32\textwidth}
  \includegraphics[width=\linewidth]{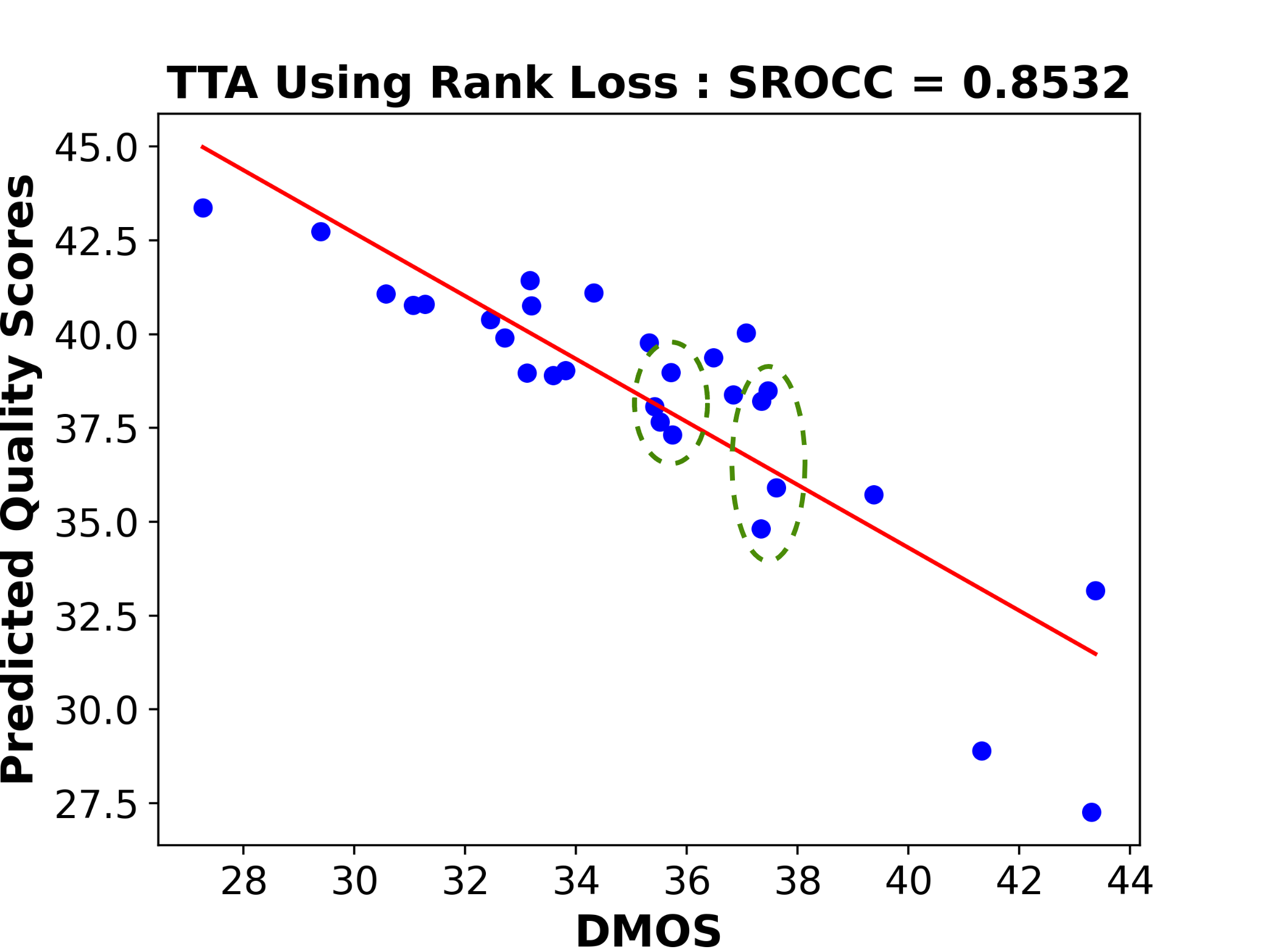}
\endminipage\hfill
\minipage{0.32\textwidth}%
  \includegraphics[width=\linewidth]{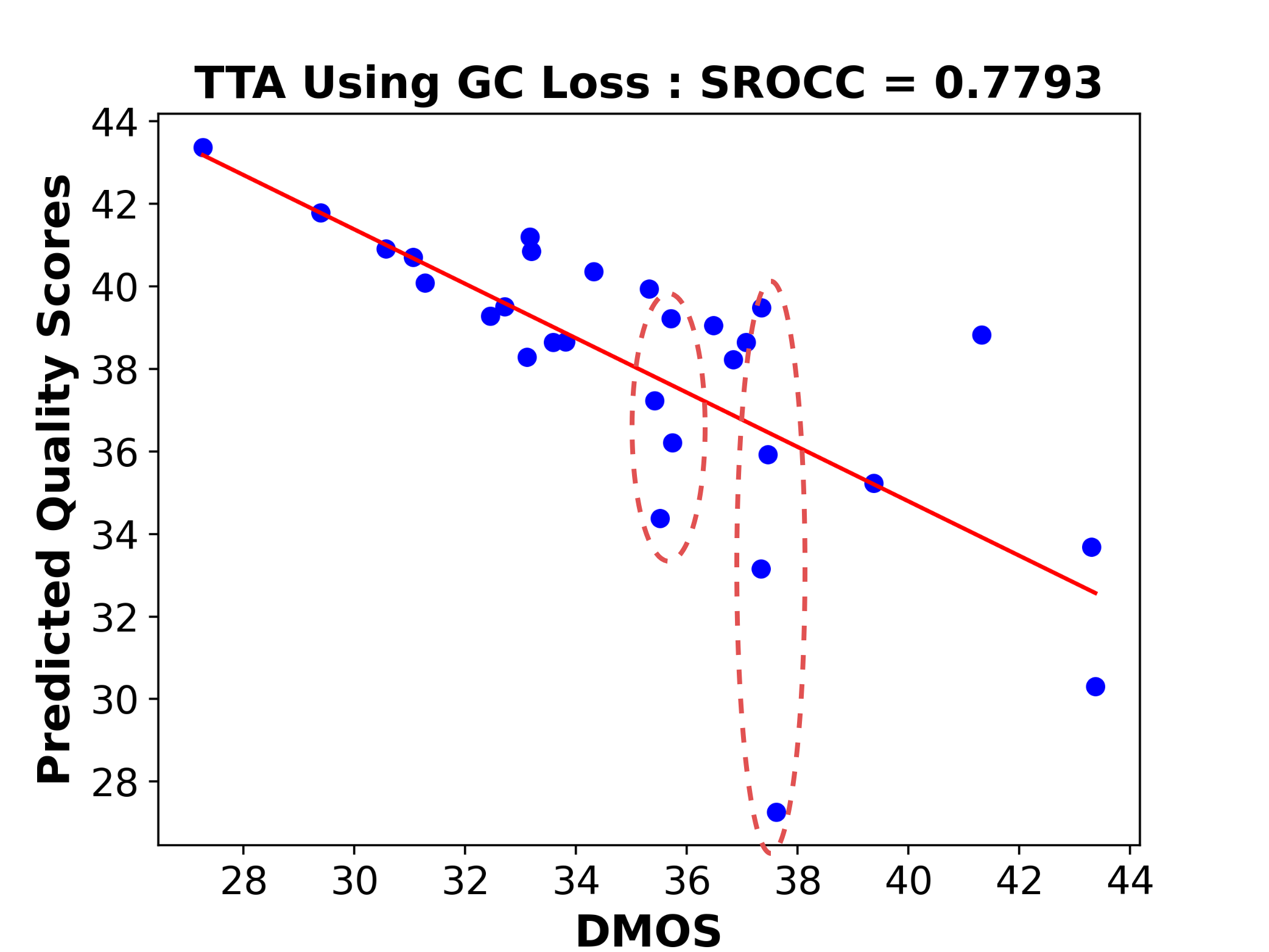}
\endminipage
\caption{Scatter plot for predicting quality of similar quality images from LIVE-IQA dataset for the source model and TTA using only rank loss or GC loss. We mark two sets of images having similar human opinion score (DMOS) by ellipses in each case.}
\label{scatter}
\end{figure*}

We also observe from Table \ref{t1} that our approach outperforms the rotation prediction task in most cases. Other TTA ideas, such as TENT \cite{c13} and training using masked autoencoders \cite{mae} are not applicable for the IQA task. In particular, the notion of entropy in regression tasks is not clear. On the other hand, masked reconstructions tend to lead to quality degradation. Thus, we do not compare with them.

\subsection{Ablation Study \& Other Experiments} \label{ablation}

\textbf{Need for Rank Loss and GC Loss for TTA.} 
We perform an ablation study with respect to the two losses in Table \ref{t2}. We see from the results that the rank loss and the GC loss individually always improve on the source models. Further, the combination of the rank and GC loss provides an even better performance over the individual losses in most evaluation scenarios. Even in scenarios, where the combination loss achieves the second-best performance, its performance is very close to the best. For the rest of the experiments in this section, we present results on all four datasets using the TReS method alone.

\begin{figure}[h]
  \centering
  \includegraphics[width=\columnwidth]{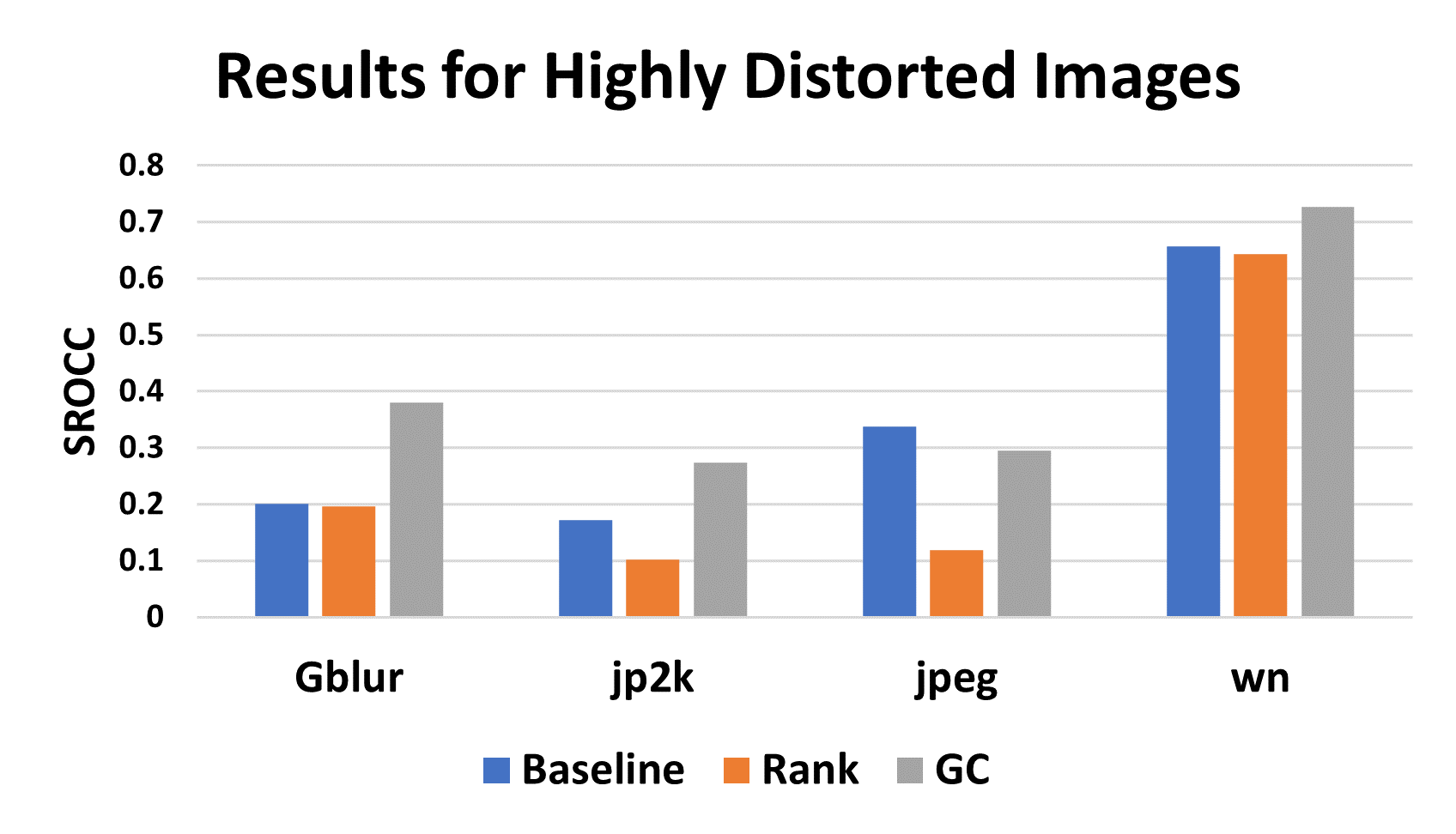}
\caption{TTA-IQA on different distortion-specific images from LIVE-IQA dataset with respect to individual losses}
\label{high_rank}
\end{figure}

We now discuss scenarios where the GC loss is particularly more useful than the rank loss. If the test image is extremely distorted, none of the three kinds of distortion types can create much difference in the perceptual quality of the two degraded versions. To understand this further, we consider highly distorted images (corresponding to a differential mean opinion score (DMOS) greater than 70) of different distortion types in the LIVE-IQA dataset. We apply TTA only for these images using different losses. The SROCC performance for these images in Figure \ref{high_rank} reveal that the rank loss is not very effective. However, the GC loss works well in such scenarios. 

Conversely, we also discuss scenarios where the rank loss is more useful than the GC loss. To illustrate this idea,  we select multiple images having similar quality with DMOS in $[28,44]$ from the LIVE-IQA dataset. We adapt the source model for this set of images using both the GC and rank losses. From Figure \ref{scatter}, we observe that the rank loss leads to much better adaptation in terms of the resulting model correlating with the ground truth scores. Intuitively, when the input images have a similar quality, the pseudo-labels given by the source model may be noisy, leading to an inaccurate grouping of the images into the lower and higher quality groups. This can adversely affect the GC loss. So the rank loss is more effective than the GC loss for test batches with similar quality images.

\textbf{Effect of Selecting Multiple Distortion Types in the Rank Loss.} 
We experiment with the impact of different kinds of distortion types while incorporating the rank loss. In particular, we consider three scenarios: one where only one of the distortion types is used in the rank loss, another where all the three types are used to obtain three loss terms which are summed together to update the model, and a third, where only one of the distortion types is chosen based on the discussion in Section \ref{rank}. We observe from Table \ref{mul} that choosing the distortion type with the maximal difference in quality gives the best performance for the rank loss. This experiment validates our hypothesis that only a rank loss that can discriminate between the degraded image pairs is useful for TTA.

\begin{table}[h]
\centering
\begin{tabular}{|c|c|c|c|c|c|}
\hline
Database  & Blur & Comp & Noise  & All  & Best \\ \hline
KonIQ-10k & 0.656 & 0.656 & \textbf{0.667} & 0.615 & 0.656 \\ 
PIPAL     & 0.384 & 0.415 & 0.389 & 0.415 & \textbf{0.417} \\ 
CID2013   & 0.527 & 0.609 & 0.606 & 0.560 & \textbf{0.609} \\ 
LIVE-IQA  & 0.578 & 0.605 & 0.618 & 0.634 & \textbf{0.671} \\ \hline
\end{tabular}
\vspace{4pt}
\caption{Impact of selecting multiple distortion types on rank loss}
\label{mul}
\end{table}

\textbf{Selection of Group Size by Varying $p$.} We explore different values of $p$ for constructing the GC loss. For a batch size of 8, possible choices of the group size are $2,3,4$ corresponding to $p=0.25,0.375,0.5$ respectively. From Table \ref{t4}, we observe that the performances are roughly similar for different values of $p$ with a slightly superior performance for $p=0.25$. We note that as $p$ increases, the groups are larger and probably less discriminative in terms of quality, leading to a slightly poorer performance.

\begin{table}[h]
\centering
\begin{tabular}{|c|c|c|c|}
\hline
Database  & $p$=0.25  & $p$=0.375 & $p$=0.5   \\ \hline
KonIQ-10k & \textbf{0.6516}  &  0.6497       & 0.6467        \\
PIPAL     & \textbf{0.4666} & 0.4602 & 0.4616 \\
CID2013   & 0.5366 & 0.5486 & \textbf{0.5536} \\
LIVE-IQA  & 0.7160 & \textbf{0.7171} & 0.6937 \\ \hline
\end{tabular}
\vspace{4pt}
\caption{Impact of varying group size using $p$ for GC loss}
\label{t4}
\end{table}

\textbf{Selection of Number of Groups for GC Loss.} In our formulation, we define the GC loss only between  two contrastive groups. In principle, we can extend our idea of GC loss to multiple groups. In particular, we can cluster the images in a batch into multiple groups and apply the GC loss between every pair of groups and sum the loss terms. Thus, images coming from the same group act as positive pairs, and images from two different groups are considered as negative pairs. From Table \ref{t5}, we observe that the number of groups $G$ does not impact the performance much.

\begin{table}[h]
\centering
\begin{tabular}{|c|c|c|c|}
\hline
Database  & $G=2$ & $G=3$ & $G=4$ \\ \hline
KonIQ-10k & 0.6516     & \textbf{0.6524}     & 0.6509     \\
PIPAL     & 0.4666 & \textbf{0.4668} & 0.4663 \\
CID2013   & 0.5366 & 0.5327 & \textbf{0.5388}  \\
LIVE-IQA  & \textbf{0.7160} & 0.7011 & 0.6883 \\ \hline
\end{tabular}
\vspace{4pt}
\caption{Impact of varying number of groups $G$ for GC loss}
\label{t5}
\end{table}

\begin{figure}[h]
\begin{center}
\includegraphics[width=0.9\columnwidth]{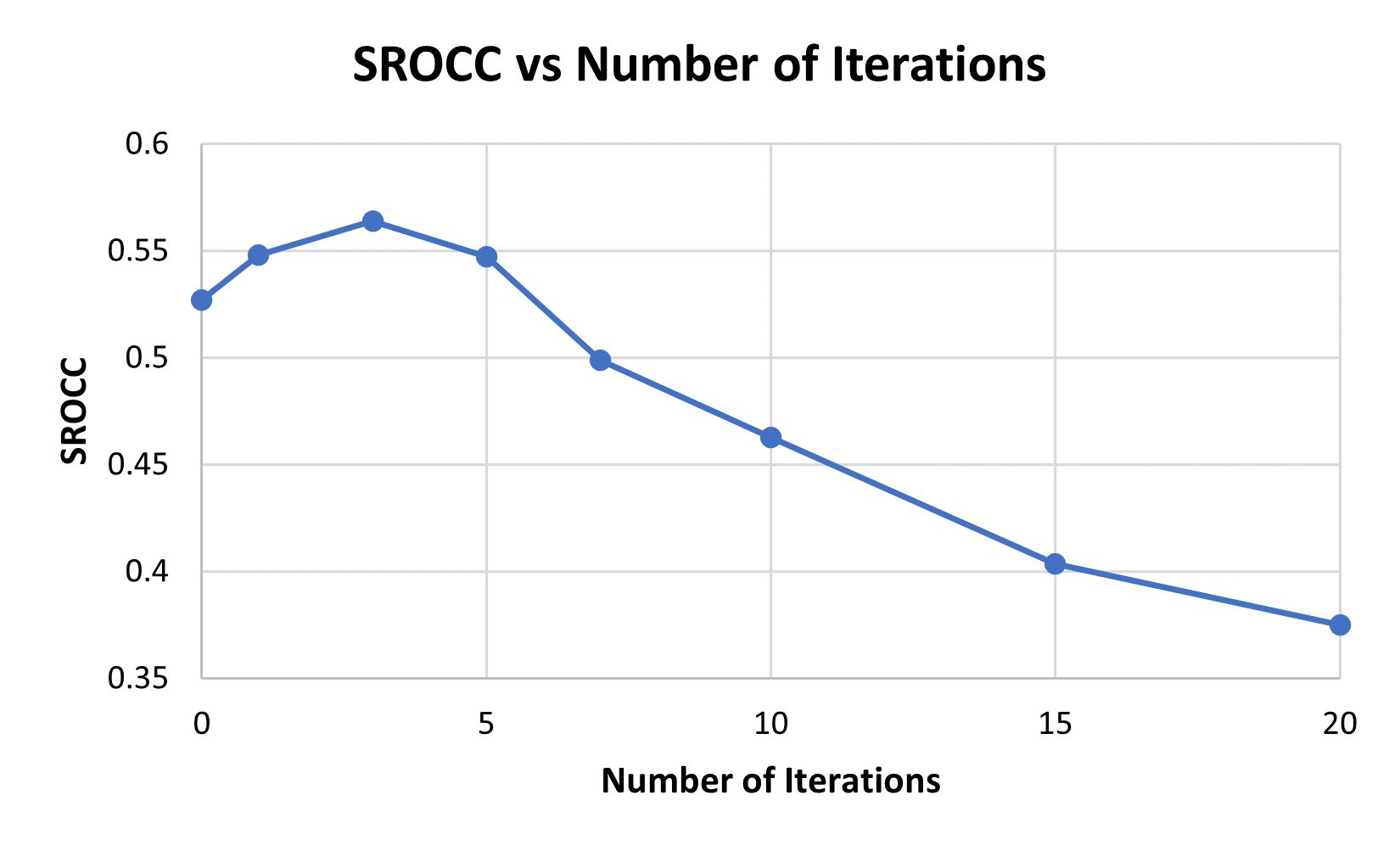}
\caption{Effect of increasing number of iterations}
\vspace{-15pt}
\label{fig6}
\end{center}
\end{figure}

\textbf{Choice of Number of Iterations for Learning Auxiliary Task.} We also examine the effect of the number of iterations of parameter updates during TTA. Figure \ref{fig6} shows that the performance on CID2013 dataset is fairly robust until 4 iterations. Beyond that, the model overfits the auxiliary task and leads to poor performance at test time. 

\section{Conclusion}
Test time adaptation has become very popular due to its simplicity and the lack of need for end-to-end training. Our work is perhaps one of the first attempts to design the method of TTA in the context of blind IQA. While most IQA methods focus on making the models robust enough to perform well for cross-database experiments, our TTA-IQA method can outperform existing state-of-the-art methods because of the adaptation at test time. We formulate novel self-supervised auxiliary tasks using the rank and group contrastive losses,  which can learn quality-aware information from the test data. While primarily explored TTA for IQA, it would be interesting to understand the role of TTA for video quality assessment as well. 

\noindent\textbf{Acknowledgement:} This work was supported in part by a grant from the Department of Science and Technology, Government of India, under grant CRG/2020/003516.

\appendix
\begin{center}
    \section*{\Large Appendix}
\end{center}

\section{Experiments with Transformer Parameters Adaptation}
We evaluate the performance of TTA-IQA by updating transformer parameters for better adaptation. For TReS \cite{tres} and MUSIQ \cite{c67}, we incorporate the transformer as a part of feature extractor. Thus, only the last fully connected (FC) layer works as the quality regressor. Current literature \cite{tta-tf} shows that layer normalization (LN) parameters of transformers are a good choice for test time adaptation. In the case of vision transformer, the CLS token is also used for adaptation. Table \ref{tf} shows the performance on all four datasets by optimizing various parameters using the combination of rank and group contrastive loss. We observe that adaptation of transformer parameters alone gives a performance equivalent to the adaptation of the batch normalization (BN) parameters of convolutional neural network (CNN) backbone. 
Thus, it is possible to update models that only use a transformer and achieve significant gains using TTA. 

\begin{table*}[t]
\centering
\begin{tabular}{|c|c|cc|cc|cc|cc|}
\hline
\multirow{2}{*}{Backbone} & Database  & \multicolumn{2}{c|}{KONIQ} & \multicolumn{2}{c|}{PIPAL} & \multicolumn{2}{c|}{CID2013} & \multicolumn{2}{c|}{LIVE-IQA} \\ \cline{2-10} 
                          & Method    & SRCC         & PLCC        & SRCC         & PLCC        & SRCC          & PLCC         & SRCC          & PLCC          \\ \hline
\multirow{4}{*}{TReS}      & Baseline & 0.6520          & 0.6955           & 0.3845          & 0.4078          & 0.5272          & 0.6463            & 0.5435          & 0.4450          \\
                          & BN Only   & 0.6731      & 0.7151      & 0.4392       & 0.2710       & 0.6173        & 0.6800         & 0.6707        & 0.6006    \\
                          & LN Only   & 0.6694       & 0.7176     & 0.4128       & 0.2690       & 0.6193       & 0.6850      & 0.5998        & 0.5458        \\
                          & BN+LN     & 0.6621       & 0.7059      & 0.4417      & 0.3484      & 0.6123        & 0.6758       & 0.6723        & 0.5948        \\ \hline
\multirow{6}{*}{MUSIQ}    & Baseline & 0.6304         & 0.6802           & 0.3190          & 0.3414           & 0.5173          & 0.6032          & 0.2596           & 0.3351          \\
                          & BN Only   & 0.6588       & 0.7174      & 0.3772       & 0.3744      & 0.5275        & 0.6126       & 0.3350         & 0.3954        \\
                          & LN Only   & 0.6582       & 0.7155      & 0.3757       & 0.3762      & 0.5403        & 0.6109       & 0.3661        & 0.4044      \\
                          & CLS Only  & 0.6618      & 0.7207   & 0.3776       & 0.3739      & 0.5491      & 0.6212      & 0.3511        & 0.3993        \\
                          & BN+LN     & 0.6552       & 0.7145      & 0.3736       & 0.3732      & 0.5329        & 0.6095       & 0.3767        & 0.4028        \\ 
                          & BN+LN+CLS & 0.6598       & 0.7172      & 0.3751       & 0.3764    & 0.5253        & 0.6048       & 0.3569        & 0.3999        \\\hline
\end{tabular}
\vspace{4pt}
\caption{Comparison of TTA-IQA using popular transformer based NR IQA methods on authentically and synthetically distorted datasets.}
\label{tf}
\end{table*}

\begin{table*}[]
\centering
{\begin{tabular}{|c|cccc|cccc|}
\hline
Train on & \multicolumn{4}{c|}{LIVEFB}                                & \multicolumn{4}{c|}{LIVE-IQA}                                 \\ \hline
Test on  & PIPAL          & KonIQ-10k        & SPAQ       & LIVEC      & PIPAL          & CID2013        & KonIQ-10k      & LIVEC      \\ \hline
Baseline & 0.385          & 0.652          & 0.707          & 0.726          & 0.402          & 0.519          & 0.521          & 0.563         \\
TTA-IQA  & \textbf{0.428} & \textbf{0.658} & \textbf{0.755} & \textbf{0.728} & \textbf{0.449} & \textbf{0.523} & \textbf{0.522} & \textbf{0.565} \\ \hline
\end{tabular}}
\caption{SRCC performance evaluation of TTA-IQA with TReS backbone trained on LIVE-IQA database}
\label{tab:liveiqa_table}
\end{table*}

\begin{table}[]
\begin{adjustbox}{max width=\columnwidth}
\begin{tabular}{|c|c|c|c|c|}
\hline
                                                     & \multicolumn{1}{c|}{TRES} & \multicolumn{1}{c|}{MUSIQ} & \multicolumn{1}{c|}{HYPER-IQA} & \multicolumn{1}{c|}{META-IQA} \\ \hline
Baseline                                             & 0.535         & 0.404       & 0.496            & 0.591            \\
Rotation                                             & 0.529          & 0.425         & 0.479              & 0.540            \\
TTA-IQA & 0.586     & 0.450           & 0.493             & 0.608      \\ \hline
\end{tabular}
\end{adjustbox}
\caption{SRCC performance analysis of TTA-IQA on DSLR database.}
\label{tab:dslr}
\end{table}

\section{Visualizing Images that Justify Need for Both Rank and GC Loss} 
\begin{figure*}[h]
    \centering 
\begin{subfigure}{0.2\textwidth}
  \includegraphics[width=\linewidth]{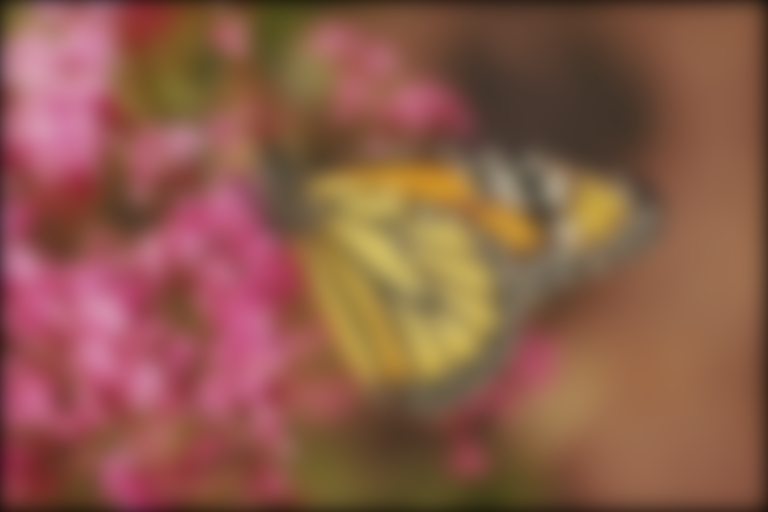}
\end{subfigure}\hfil 
\begin{subfigure}{0.2\textwidth}
  \includegraphics[width=\linewidth]{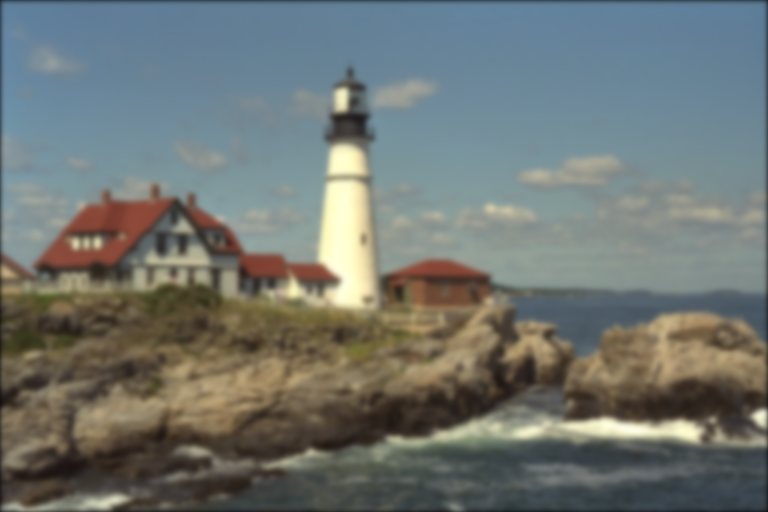}
\end{subfigure}\hfil 
\begin{subfigure}{0.2\textwidth}
  \includegraphics[width=\linewidth]{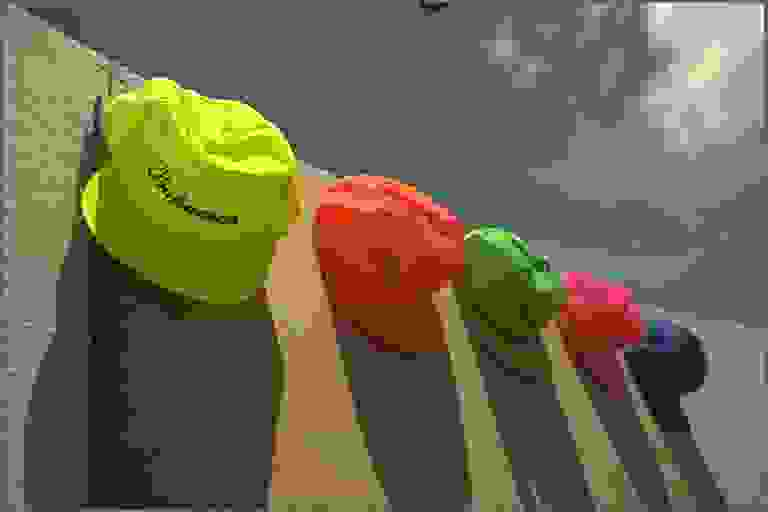}
\end{subfigure}\hfil 
\begin{subfigure}{0.2\textwidth}
  \includegraphics[width=\linewidth]{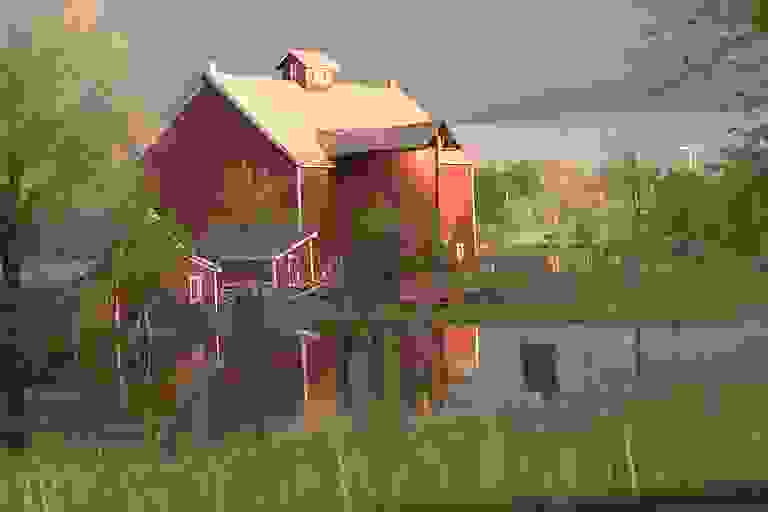}
\end{subfigure}

\medskip
\begin{subfigure}{0.2\textwidth}
  \includegraphics[width=\linewidth]{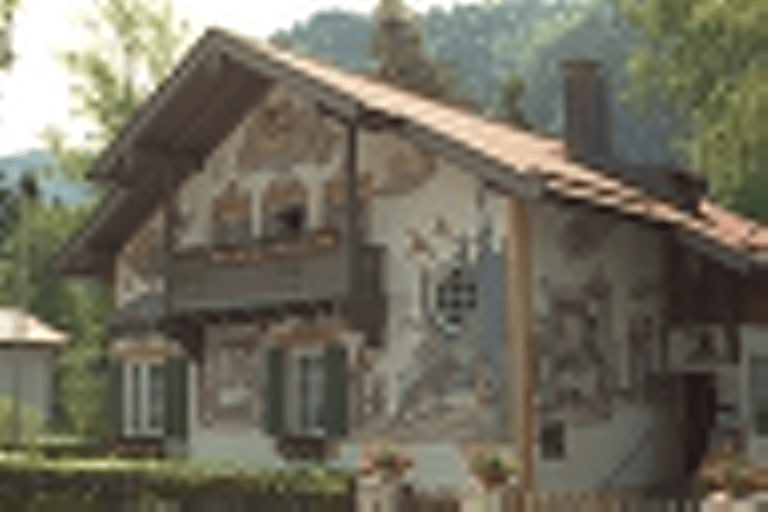}
\end{subfigure}\hfil 
\begin{subfigure}{0.2\textwidth}
  \includegraphics[width=\linewidth]{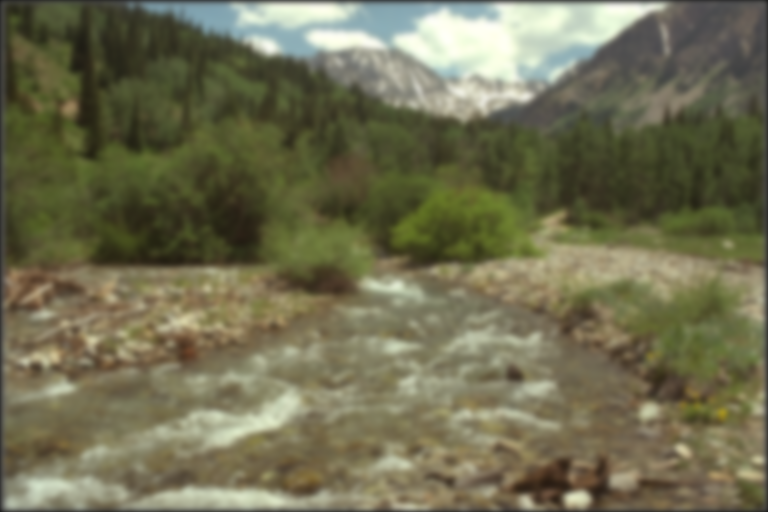}
\end{subfigure}\hfil 
\begin{subfigure}{0.2\textwidth}
  \includegraphics[width=\linewidth]{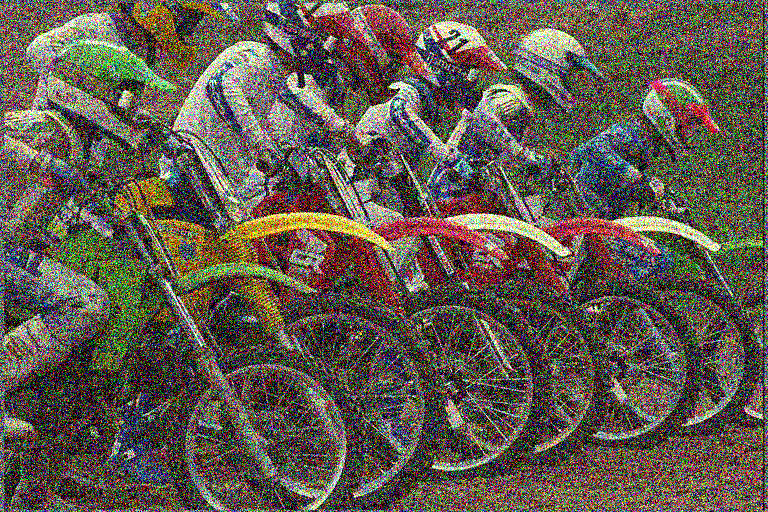}
\end{subfigure}\hfil 
\begin{subfigure}{0.2\textwidth}
  \includegraphics[width=\linewidth]{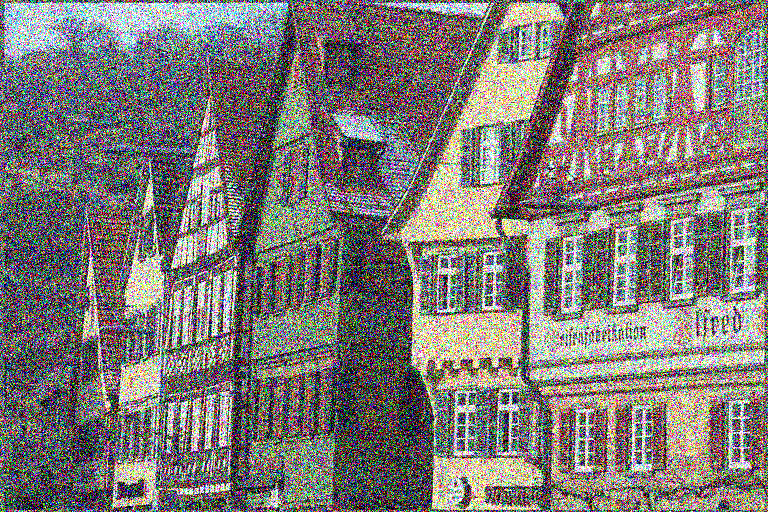}
\end{subfigure}
\caption{Examples of highly distorted images in which GC loss is more effective than rank loss}
\label{high_distort}
\end{figure*}

\begin{figure*}[h]
    \centering 
\begin{subfigure}{0.2\textwidth}
  \includegraphics[width=\linewidth]{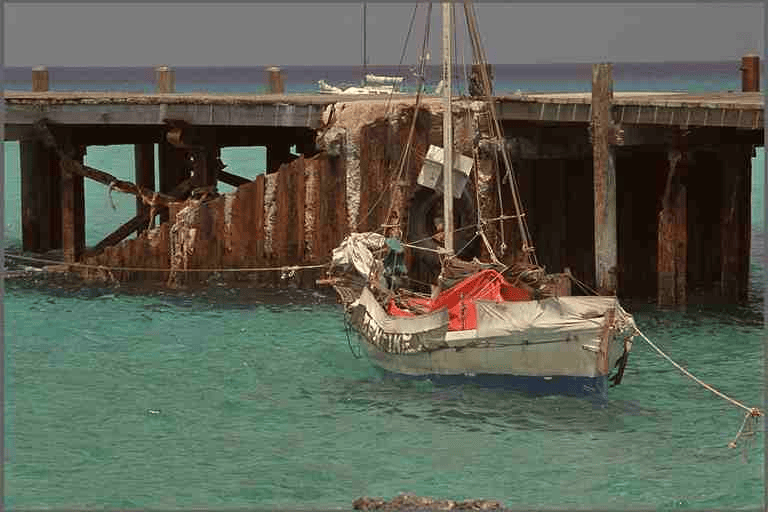}
\end{subfigure}\hfil 
\begin{subfigure}{0.2\textwidth}
  \includegraphics[width=\linewidth]{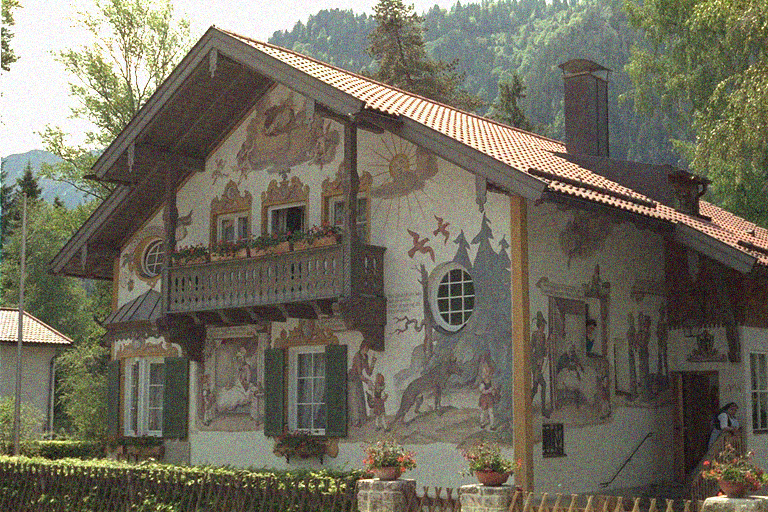}
\end{subfigure}\hfil 
\begin{subfigure}{0.2\textwidth}
  \includegraphics[width=\linewidth]{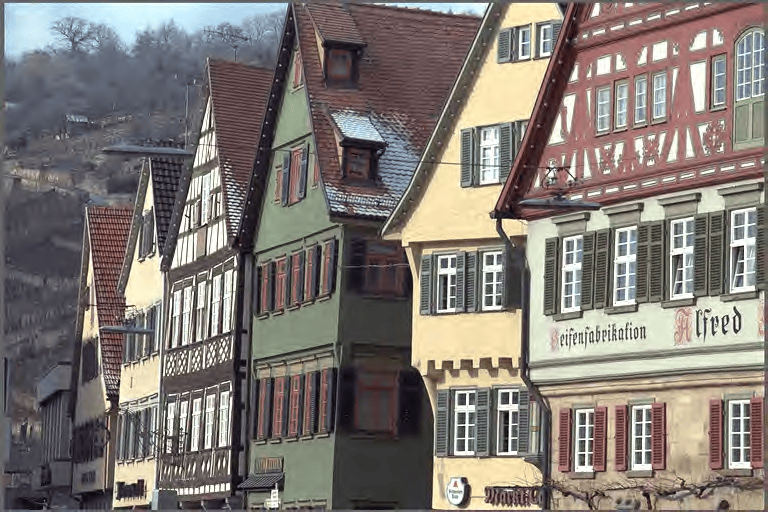}
\end{subfigure}\hfil 
\begin{subfigure}{0.2\textwidth}
  \includegraphics[width=\linewidth]{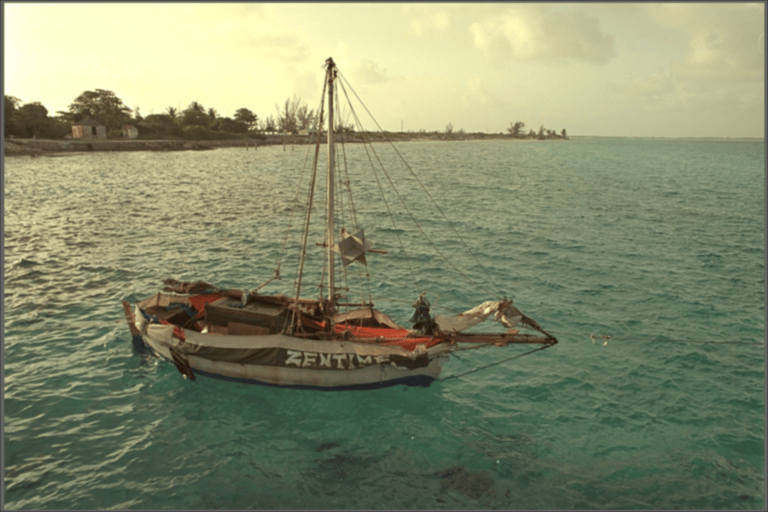}
\end{subfigure}

\medskip
\begin{subfigure}{0.2\textwidth}
  \includegraphics[width=\linewidth]{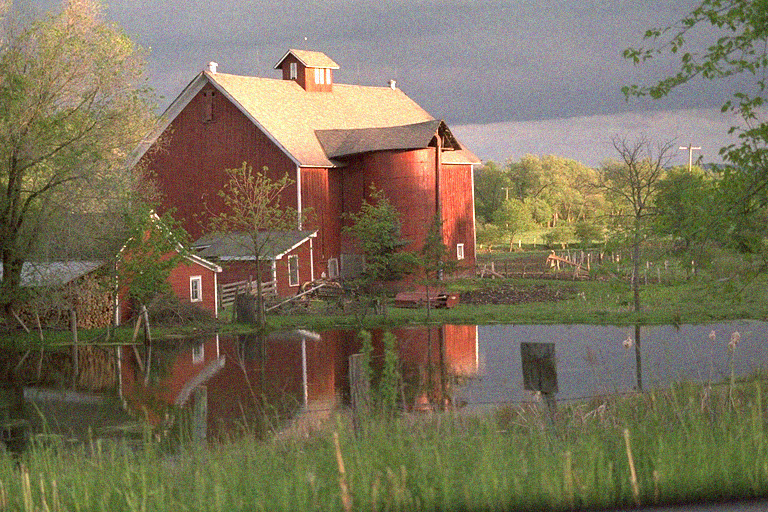}
\end{subfigure}\hfil 
\begin{subfigure}{0.2\textwidth}
  \includegraphics[width=\linewidth]{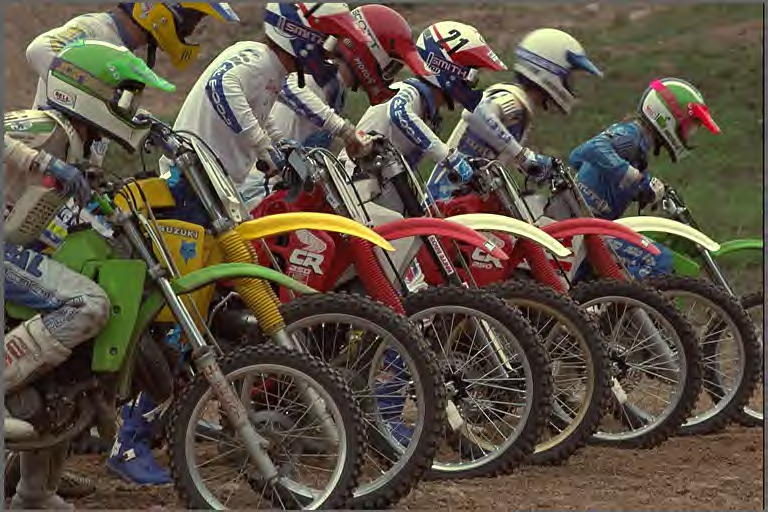}
\end{subfigure}\hfil 
\begin{subfigure}{0.2\textwidth}
  \includegraphics[width=\linewidth]{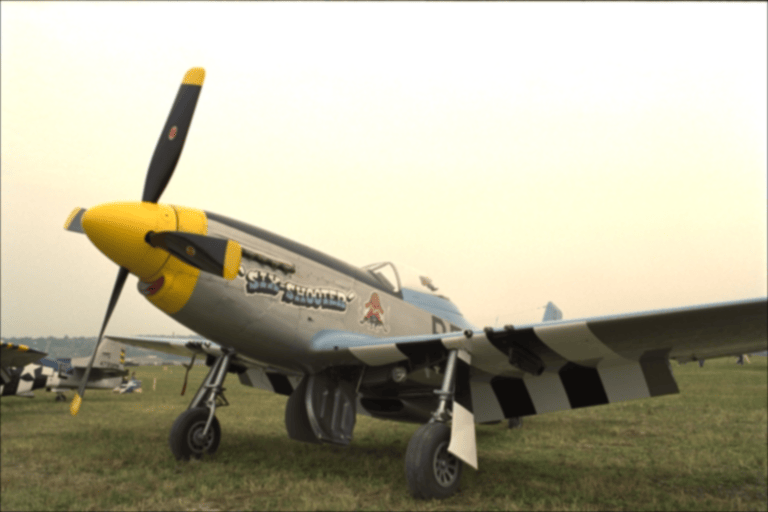}
\end{subfigure}\hfil 
\begin{subfigure}{0.2\textwidth}
  \includegraphics[width=\linewidth]{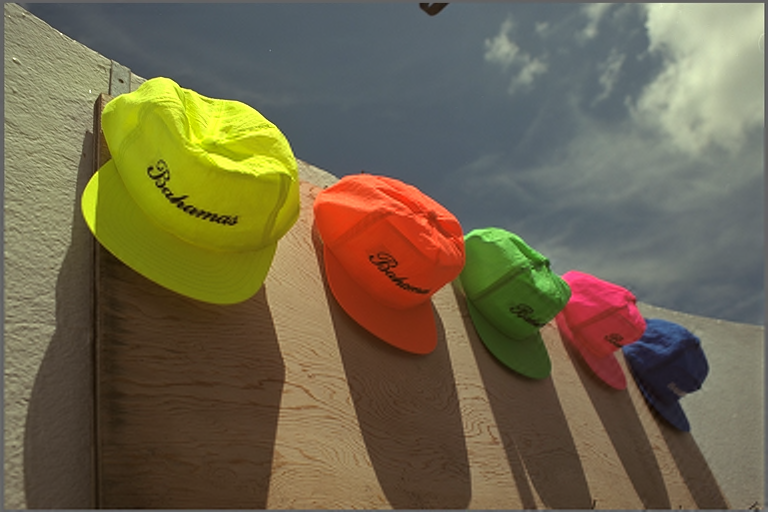}
\end{subfigure}
\caption{Examples of similar quality images in which rank loss is more effective than GC loss}
\label{similar}
\end{figure*}

In Section \ref{ablation}, we justify the need for both the rank and GC loss for effective TTA. Here we give a few visual examples of images corresponding to that analysis. In Figure \ref{high_distort}, we observe that the images have very poor quality. Hence, distorting these images further creates distorted versions that have perceptually indistinguishable quality ratings. On the other hand, Figure \ref{similar} shows similar quality images. Here, as the images have almost similar visual quality, it is difficult to form two different quality groups based on pseudo-labels given by the source model. 

\section{Performance of TTA-IQA on Other Databases}
 
\subsection{Performance evaluation with synthetic database as source database}
In the main paper, we reported performances where the source model is trained on camera captured LIVEFB \cite{c23} database and tested on various authentic and synthetic databases. In Table \ref{tab:liveiqa_table}, we provide more such evaluations with respect to different intra and inter domain comparisons. In particular, we present results when TReS \cite{tres} is trained on LIVE FB and evaluated on more intra domain datasets such as SPAQ \cite{spaq} and LIVEC \cite{clive}. We also present results when TReS is trained on a synthetic dataset such as LIVE-IQA \cite{LIVE} and tested on authentic as well as other datasets containing restored images. We observe that TTA-IQA gives a reasonable performance gain over the baseline even when there is domain shift between source (synthetic) data and target (authentic) data.

\subsection{Performance on Low-Light Restorted Database}
To understand the impact of larger domain shifts, we also evaluate on a new database DSLR \cite{llr_vignesh}, where images captured in low light are restored via various image restoration algorithms. Since novel distortions are generated while restorting such low-light images, we evaluate the performance of TTA-IQA with source database as LIVEFB and target datasbase as DSLR database. We see that TTA-IQA helps improve the performance of most of the methods. 

{\small
\bibliographystyle{ieee_fullname}
\bibliography{paper}

\begin{thebibliography}{10}\itemsep=-1pt

\bibitem{c48}
Piotr Bojanowski and Armand Joulin.
\newblock Unsupervised learning by predicting noise.
\newblock In Doina Precup and Yee~Whye Teh, editors, {\em Proceedings of the
  34th International Conference on Machine Learning}, volume~70 of {\em
  Proceedings of Machine Learning Research}, pages 517--526. PMLR, 06--11 Aug
  2017.

\bibitem{nrfr}
Sebastian Bosse, Dominique Maniry, Klaus-Robert Müller, Thomas Wiegand, and
  Wojciech Samek.
\newblock Deep neural networks for no-reference and full-reference image
  quality assessment.
\newblock {\em IEEE Transactions on Image Processing}, 27(1):206--219, 2018.

\bibitem{c49}
Mathilde Caron, Piotr Bojanowski, Armand Joulin, and Matthijs Douze.
\newblock Deep clustering for unsupervised learning of visual features.
\newblock In Vittorio Ferrari, Martial Hebert, Cristian Sminchisescu, and Yair
  Weiss, editors, {\em Computer Vision -- ECCV 2018}, pages 139--156, Cham,
  2018. Springer International Publishing.

\bibitem{simclr}
Ting Chen, Simon Kornblith, Mohammad Norouzi, and Geoffrey Hinton.
\newblock A simple framework for contrastive learning of visual
  representations.
\newblock In {\em Proceedings of the 37th International Conference on Machine
  Learning}, ICML'20. JMLR.org, 2020.

\bibitem{c45}
Carl Doersch, Abhinav Gupta, and Alexei~A. Efros.
\newblock Unsupervised visual representation learning by context prediction.
\newblock In {\em Proceedings of the IEEE International Conference on Computer
  Vision (ICCV)}, December 2015.

\bibitem{spaq}
Yuming Fang, Hanwei Zhu, Yan Zeng, Kede Ma, and Zhou Wang.
\newblock Perceptual quality assessment of smartphone photography.
\newblock In {\em 2020 IEEE/CVF Conference on Computer Vision and Pattern
  Recognition (CVPR)}, pages 3674--3683, 2020.

\bibitem{mae}
Yossi Gandelsman, Yu Sun, Xinlei Chen, and Alexei~A. Efros.
\newblock Test-time training with masked autoencoders.
\newblock {\em arXiv preprint arXiv:2209.07522}, 2022.

\bibitem{clive}
Deepti Ghadiyaram and Alan~C. Bovik.
\newblock Massive online crowdsourced study of subjective and objective picture
  quality.
\newblock {\em IEEE Transactions on Image Processing}, 25(1):372--387, 2016.

\bibitem{tres}
S~Alireza Golestaneh, Saba Dadsetan, and Kris~M Kitani.
\newblock No-reference image quality assessment via transformers, relative
  ranking, and self-consistency.
\newblock In {\em Proceedings of the IEEE/CVF Winter Conference on Applications
  of Computer Vision}, pages 3209--3218, 2022.

\bibitem{c81}
Dan Hendrycks and Thomas Dietterich.
\newblock Benchmarking neural network robustness to common corruptions and
  perturbations.
\newblock {\em Proceedings of the International Conference on Learning
  Representations}, 2019.

\bibitem{c24}
Vlad Hosu, Hanhe Lin, Tamas Sziranyi, and Dietmar Saupe.
\newblock Koniq-10k: An ecologically valid database for deep learning of blind
  image quality assessment.
\newblock {\em Trans. Img. Proc.}, 29:4041–4056, jan 2020.

\bibitem{tta-proto}
Yusuke Iwasawa and Yutaka Matsuo.
\newblock Test-time classifier adjustment module for model-agnostic domain
  generalization.
\newblock In M. Ranzato, A. Beygelzimer, Y. Dauphin, P.S. Liang, and J.~Wortman
  Vaughan, editors, {\em Advances in Neural Information Processing Systems},
  volume~34, pages 2427--2440. Curran Associates, Inc., 2021.

\bibitem{c25}
Gu Jinjin, Cai Haoming, Chen Haoyu, Ye Xiaoxing, Jimmy~S. Ren, and Dong Chao.
\newblock Pipal: A large-scale image quality assessment dataset for perceptual
  image restoration.
\newblock In Andrea Vedaldi, Horst Bischof, Thomas Brox, and Jan-Michael Frahm,
  editors, {\em Computer Vision -- ECCV 2020}, pages 633--651, Cham, 2020.
  Springer International Publishing.

\bibitem{llr_vignesh}
Vignesh Kannan, Sameer Malik, Nithin~C. Babu, and Rajiv Soundararajan.
\newblock Quality assessment of low-light restored images: A subjective study
  and an unsupervised model.
\newblock {\em IEEE Access}, 11:68216--68230, 2023.

\bibitem{c67}
Junjie Ke, Qifei Wang, Yilin Wang, Peyman Milanfar, and Feng Yang.
\newblock Musiq: Multi-scale image quality transformer.
\newblock In {\em Proceedings of the IEEE/CVF International Conference on
  Computer Vision (ICCV)}, pages 5148--5157, October 2021.

\bibitem{c72}
Diederik~P Kingma and Jimmy Ba.
\newblock Adam: A method for stochastic optimization.
\newblock {\em arXiv preprint arXiv:1412.6980}, 2014.

\bibitem{tta-tf}
Takeshi Kojima, Yutaka Matsuo, and Yusuke Iwasawa.
\newblock Robustifying vision transformer without retraining from scratch by
  test-time class-conditional feature alignment.
\newblock In Lud~De Raedt, editor, {\em Proceedings of the Thirty-First
  International Joint Conference on Artificial Intelligence, {IJCAI-22}}, pages
  1009--1016. International Joint Conferences on Artificial Intelligence
  Organization, 7 2022.
\newblock Main Track.

\bibitem{c77}
Nikos Komodakis and Spyros Gidaris.
\newblock {Unsupervised representation learning by predicting image rotations}.
\newblock In {\em {International Conference on Learning Representations
  (ICLR)}}, Vancouver, Canada, Apr. 2018.

\bibitem{c47}
Gustav Larsson, Michael Maire, and Gregory Shakhnarovich.
\newblock Colorization as a proxy task for visual understanding.
\newblock In {\em Proceedings of the IEEE Conference on Computer Vision and
  Pattern Recognition (CVPR)}, July 2017.

\bibitem{c40}
Jian Liang, Dapeng Hu, and Jiashi Feng.
\newblock Do we really need to access the source data? {S}ource hypothesis
  transfer for unsupervised domain adaptation.
\newblock In Hal~Daumé III and Aarti Singh, editors, {\em Proceedings of the
  37th International Conference on Machine Learning}, volume 119 of {\em
  Proceedings of Machine Learning Research}, pages 6028--6039. PMLR, 13--18 Jul
  2020.

\bibitem{kadid}
Hanhe Lin, Vlad Hosu, and Dietmar Saupe.
\newblock Kadid-10k: A large-scale artificially distorted iqa database.
\newblock In {\em 2019 Eleventh International Conference on Quality of
  Multimedia Experience (QoMEX)}, pages 1--3, 2019.

\bibitem{c57}
Xialei Liu, Joost van~de Weijer, and Andrew~D. Bagdanov.
\newblock Rankiqa: Learning from rankings for no-reference image quality
  assessment.
\newblock In {\em Proceedings of the IEEE International Conference on Computer
  Vision (ICCV)}, Oct 2017.

\bibitem{c19}
Yutao Liu, Ke Gu, Yongbing Zhang, Xiu Li, Guangtao Zhai, Debin Zhao, and Wen
  Gao.
\newblock Unsupervised blind image quality evaluation via statistical
  measurements of structure, naturalness, and perception.
\newblock {\em IEEE Transactions on Circuits and Systems for Video Technology},
  30(4):929--943, 2020.

\bibitem{c37}
Yuejiang Liu, Parth Kothari, Bastien van Delft, Baptiste Bellot-Gurlet, Taylor
  Mordan, and Alexandre Alahi.
\newblock Ttt++: When does self-supervised test-time training fail or thrive?
\newblock In M. Ranzato, A. Beygelzimer, Y. Dauphin, P.S. Liang, and J.~Wortman
  Vaughan, editors, {\em Advances in Neural Information Processing Systems},
  volume~34, pages 21808--21820. Curran Associates, Inc., 2021.

\bibitem{c58}
Kede Ma, Wentao Liu, Kai Zhang, Zhengfang Duanmu, Zhou Wang, and Wangmeng Zuo.
\newblock End-to-end blind image quality assessment using deep neural networks.
\newblock {\em IEEE Transactions on Image Processing}, 27(3):1202--1213, 2018.

\bibitem{c16}
Anish Mittal, Anush~Krishna Moorthy, and Alan~Conrad Bovik.
\newblock No-reference image quality assessment in the spatial domain.
\newblock {\em IEEE Transactions on Image Processing}, 21(12):4695--4708, 2012.

\bibitem{c14}
Anush~Krishna Moorthy and Alan~Conrad Bovik.
\newblock Blind image quality assessment: From natural scene statistics to
  perceptual quality.
\newblock {\em IEEE Transactions on Image Processing}, 20(12):3350--3364, 2011.

\bibitem{tta-conf}
Chaithanya~Kumar Mummadi, Robin Hutmacher, Kilian Rambach, Evgeny Levinkov,
  Thomas Brox, and Jan~Hendrik Metzen.
\newblock Test-time adaptation to distribution shift by confidence maximization
  and input transformation, 2021.

\bibitem{c74}
Zachary Nado, Shreyas Padhy, D Sculley, Alexander D'Amour, Balaji
  Lakshminarayanan, and Jasper Snoek.
\newblock Evaluating prediction-time batch normalization for robustness under
  covariate shift.
\newblock {\em arXiv preprint arXiv:2006.10963}, 2020.

\bibitem{tid}
Nikolay Ponomarenko, Lina Jin, Oleg Ieremeiev, Vladimir Lukin, Karen
  Egiazarian, Jaakko Astola, Benoit Vozel, Kacem Chehdi, Marco Carli, Federica
  Battisti, and C.-C. {Jay Kuo}.
\newblock Image database tid2013: Peculiarities, results and perspectives.
\newblock {\em Signal Processing: Image Communication}, 30:57--77, 2015.

\bibitem{saad}
Michele~A. Saad, Alan~C. Bovik, and Christophe Charrier.
\newblock Blind image quality assessment: A natural scene statistics approach
  in the dct domain.
\newblock {\em IEEE Transactions on Image Processing}, 21(8):3339--3352, 2012.

\bibitem{c39}
Steffen Schneider, Evgenia Rusak, Luisa Eck, Oliver Bringmann, Wieland Brendel,
  and Matthias Bethge.
\newblock Improving robustness against common corruptions by covariate shift
  adaptation.
\newblock In H. Larochelle, M. Ranzato, R. Hadsell, M.F. Balcan, and H. Lin,
  editors, {\em Advances in Neural Information Processing Systems}, volume~33,
  pages 11539--11551. Curran Associates, Inc., 2020.

\bibitem{LIVE}
H.R. Sheikh, M.F. Sabir, and A.C. Bovik.
\newblock A statistical evaluation of recent full reference image quality
  assessment algorithms.
\newblock {\em IEEE Transactions on Image Processing}, 15(11):3440--3451, 2006.

\bibitem{c63}
Shaolin Su, Qingsen Yan, Yu Zhu, Cheng Zhang, Xin Ge, Jinqiu Sun, and Yanning
  Zhang.
\newblock Blindly assess image quality in the wild guided by a self-adaptive
  hyper network.
\newblock In {\em Proceedings of the IEEE/CVF Conference on Computer Vision and
  Pattern Recognition (CVPR)}, June 2020.

\bibitem{ttt-rot}
Yu Sun, Xiaolong Wang, Zhuang Liu, John Miller, Alexei Efros, and Moritz Hardt.
\newblock Test-time training with self-supervision for generalization under
  distribution shifts.
\newblock In Hal~Daumé III and Aarti Singh, editors, {\em Proceedings of the
  37th International Conference on Machine Learning}, volume 119 of {\em
  Proceedings of Machine Learning Research}, pages 9229--9248. PMLR, 13--18 Jul
  2020.

\bibitem{c26}
Toni Virtanen, Mikko Nuutinen, Mikko Vaahteranoksa, Pirkko Oittinen, and Jukka
  Häkkinen.
\newblock Cid2013: A database for evaluating no-reference image quality
  assessment algorithms.
\newblock {\em IEEE Transactions on Image Processing}, 24(1):390--402, 2015.

\bibitem{c13}
Dequan Wang, Evan Shelhamer, Shaoteng Liu, Bruno Olshausen, and Trevor Darrell.
\newblock Tent: Fully test-time adaptation by entropy minimization.
\newblock In {\em International Conference on Learning Representations}, 2021.

\bibitem{c53}
Jingtao Xu, Peng Ye, Qiaohong Li, Haiqing Du, Yong Liu, and David Doermann.
\newblock Blind image quality assessment based on high order statistics
  aggregation.
\newblock {\em IEEE Transactions on Image Processing}, 25(9):4444--4457, 2016.

\bibitem{c52}
Peng Ye, Jayant Kumar, Le Kang, and David Doermann.
\newblock Unsupervised feature learning framework for no-reference image
  quality assessment.
\newblock In {\em 2012 IEEE conference on computer vision and pattern
  recognition}, pages 1098--1105. IEEE, 2012.

\bibitem{c55}
Peng Ye, Jayant Kumar, Le Kang, and David Doermann.
\newblock Unsupervised feature learning framework for no-reference image
  quality assessment.
\newblock In {\em 2012 IEEE Conference on Computer Vision and Pattern
  Recognition}, pages 1098--1105, 2012.

\bibitem{c23}
Zhenqiang Ying, Haoran Niu, Praful Gupta, Dhruv Mahajan, Deepti Ghadiyaram, and
  Alan Bovik.
\newblock From patches to pictures (paq-2-piq): Mapping the perceptual space of
  picture quality.
\newblock In {\em Proceedings of the IEEE/CVF Conference on Computer Vision and
  Pattern Recognition (CVPR)}, June 2020.

\bibitem{c18}
Guangtao Zhai, Xiongkuo Min, and Ning Liu.
\newblock Free-energy principle inspired visual quality assessment: An
  overview.
\newblock {\em Digital Signal Processing}, 91:11--20, 2019.

\bibitem{c61}
Weixia Zhang, Kede Ma, Jia Yan, Dexiang Deng, and Zhou Wang.
\newblock Blind image quality assessment using a deep bilinear convolutional
  neural network.
\newblock {\em IEEE Transactions on Circuits and Systems for Video Technology},
  30(1):36--47, 2020.

\bibitem{meta}
Hancheng Zhu, Leida Li, Jinjian Wu, Weisheng Dong, and Guangming Shi.
\newblock Metaiqa: Deep meta-learning for no-reference image quality
  assessment.
\newblock In {\em Proceedings of the IEEE/CVF Conference on Computer Vision and
  Pattern Recognition (CVPR)}, June 2020.

\end{thebibliography}
}

\end{document}